\documentclass[11pt]{article}
\usepackage{acl}
\usepackage{times}
\usepackage{latexsym}
\usepackage[utf8]{inputenc}
\usepackage{microtype}
\usepackage{inconsolata}
\usepackage{amsmath,amssymb,bm}
\usepackage{tikz}
\providecommand{\Description}[1]{}
\usepackage{placeins}   % Keep wide result tables before the reference list

\usepackage{booktabs,tabularx}
\usepackage{multirow}
\usepackage{colortbl}
\definecolor{resultlavender}{HTML}{F3E8F7}
\usepackage{graphicx}   % 提供 \resizebox
\usepackage{subcaption} % 提供 subfigure（fig5 四合一子图 a/b/c/d）
\usepackage{arydshln}   % 提供 \hdashline 水平虚线（国产/国外分隔）
\usepackage{enumitem}   % 提供 itemize 的 [nosep] 等紧凑间距选项
\usepackage{algorithm}      % FA-GRPO 伪代码块（浮动环境）
\usepackage{algpseudocode}  % algorithmicx 伪代码排版

\usepackage{xcolor}         % Figure 6 failure-case cards
\usepackage[skins]{tcolorbox}          % Figure 6 圆角卡片 + 淡蓝阴影
\usepackage[LY1,T1]{fontenc}
\pdfmapline{+cmcsnwa ComicSansMS <<fonts/comic.ttf}
\pdfmapline{+cmcsbwa ComicSansMS-Bold <<fonts/comicbd.ttf}
\pdfmapline{+cmcsiwa ComicSansMS-Italic <<fonts/comici.ttf}
\pdfmapline{+cmcszwa ComicSansMS-BoldItalic <<fonts/comicz.ttf}
\DeclareFontFamily{LY1}{cmcs}{}
\DeclareFontShape{LY1}{cmcs}{m}{n}{<-> cmcsnwa}{}
\DeclareFontShape{LY1}{cmcs}{b}{n}{<-> cmcsbwa}{}
\DeclareFontShape{LY1}{cmcs}{bx}{n}{<-> cmcsbwa}{}
\DeclareFontShape{LY1}{cmcs}{m}{it}{<-> cmcsiwa}{}
\DeclareFontShape{LY1}{cmcs}{b}{it}{<-> cmcszwa}{}
\DeclareFontShape{LY1}{cmcs}{bx}{it}{<-> cmcszwa}{}
\newcommand{\comicsans}{\fontencoding{LY1}\fontfamily{cmcs}\selectfont}
\DeclareTextCommandDefault{\copyright}{\textcopyright}
\usepackage{listings}       % 附录 Prompt 模板框
\lstdefinestyle{promptbox}{%
  basicstyle=\ttfamily\footnotesize,breaklines=true,breakatwhitespace=false,
  frame=single,framesep=4pt,columns=fullflexible,keepspaces=true,
  showstringspaces=false,xleftmargin=3pt,xrightmargin=3pt,aboveskip=5pt,belowskip=3pt}
\newcommand{\subcap}[1]{\par\vspace{1.5pt}{\small #1\par}\vspace{2.5pt}}
\definecolor{promptbar}{HTML}{1F2A8C}
\definecolor{promptbg}{HTML}{F5F7FE}
\newtcolorbox{promptcard}[1]{%
  enhanced, colback=promptbg, colframe=promptbar, boxrule=0.9pt,
  arc=1.5pt, outer arc=1.5pt, boxsep=0.7mm, left=4pt, right=4pt, top=1pt, bottom=1pt,
  fonttitle=\bfseries\small, coltitle=white, colbacktitle=promptbar,
  toptitle=1pt, bottomtitle=1pt, title={#1}}
\lstdefinestyle{promptin}{%
  basicstyle=\ttfamily\scriptsize,breaklines=true,breakatwhitespace=false,
  frame=none,columns=fullflexible,keepspaces=true,showstringspaces=false,
  xleftmargin=0pt,xrightmargin=0pt,aboveskip=2pt,belowskip=0pt}
\lstdefinestyle{answerbox}{%
  basicstyle=\ttfamily\fontsize{8}{9}\selectfont,breaklines=true,breakatwhitespace=false,
  literate={,}{{,\allowbreak}}1,
  frame=single,framesep=3pt,columns=fullflexible,keepspaces=true,
  showstringspaces=false,xleftmargin=3pt,xrightmargin=3pt,aboveskip=3pt,belowskip=1pt}

\allowdisplaybreaks[1]
\AtBeginDocument{%
  \setlength{\abovedisplayskip}{5pt plus 1pt minus 1pt}%
  \setlength{\belowdisplayskip}{5pt plus 1pt minus 1pt}%
  \setlength{\abovedisplayshortskip}{3pt plus 1pt minus 1pt}%
  \setlength{\belowdisplayshortskip}{3pt plus 1pt minus 1pt}%
  \setlength{\jot}{1pt}}
\newcommand{\tablefont}{\fontsize{10}{11.5}\selectfont}
\AtBeginEnvironment{table}{\tablefont}
\AtBeginEnvironment{table*}{\tablefont}
\title{PowerAtlas: Feasibility-Aware Language Models for Electricity--Computing Co-Scheduling}
\author{%
  Kaiwen Jiang\textsuperscript{1}\quad
  Siya Xu\textsuperscript{1}\thanks{Corresponding authors.}\quad
  Ziyue Zhu\textsuperscript{1}\quad
  Kaiwen Xue\textsuperscript{1}\\
  \textbf{Chao Yang\textsuperscript{2}\quad
  Anh Tuan Luu\textsuperscript{3}\quad
  Haoran Luo\textsuperscript{3,*}}\\
  \normalfont\textsuperscript{1}Beijing University of Posts and Telecommunications, Beijing, China\\
  \normalfont\textsuperscript{2}State Grid Liaoning Electric Power Co., Ltd., Liaoning, China\\
  \normalfont\textsuperscript{3}Nanyang Technological University, Singapore\\
  \normalfont\texttt{\{jiangkaiwen,xusiyaxsy,zyzhu\}@bupt.edu.cn}\\
  \normalfont\texttt{haoran.luo@ntu.edu.sg}}
\hypersetup{pdfauthor={Kaiwen Jiang, Siya Xu, Ziyue Zhu, Kaiwen Xue, Chao Yang, Anh Tuan Luu, Haoran Luo},pdftitle={PowerAtlas: Feasibility-Aware Language Models for Electricity\textendash{}Computing Co-Scheduling}}
\begin{document}
\maketitle
\suppressfloats[t] % Keep the abstract above the first figure.
\begin{abstract}
Electricity--computing co-scheduling (ECCS) decides where and when computing tasks run together with how generators are dispatched, because task placement changes nodal demand while power availability determines whether tasks can be served.
Prior work addresses such coupled problems with optimization or learned controllers, and large language models (LLMs) are increasingly used to generate structured decisions for scheduling and power systems.
Solver runtimes vary across instances, and LLM outputs that parse correctly can still violate grid and service constraints.
We formalize ECCS as a structured decision task and release ECBench, 2,000 instances with Gurobi-optimal reference schedules and executable checks for grid limits, task service, cost, and latency.
We propose PowerAtlas, which trains an LLM with supervised initialization and Feasibility-Aware GRPO on deterministic scorer feedback, and generates the joint schedule in a single response without solver calls or retrieval.
On 10--40 paired instances per backbone, selected PowerAtlas checkpoints lower violations by 89--98\% and penalized cost by up to 26$\times$ relative to base models, with residual supply deficits.
Code and data are available online.\footnote{\url{https://github.com/JAVA-Jiang/PowerAtlas}}
\end{abstract}

\section{Introduction}

\begin{figure}[!t]
  \centering
  \includegraphics[width=\columnwidth]{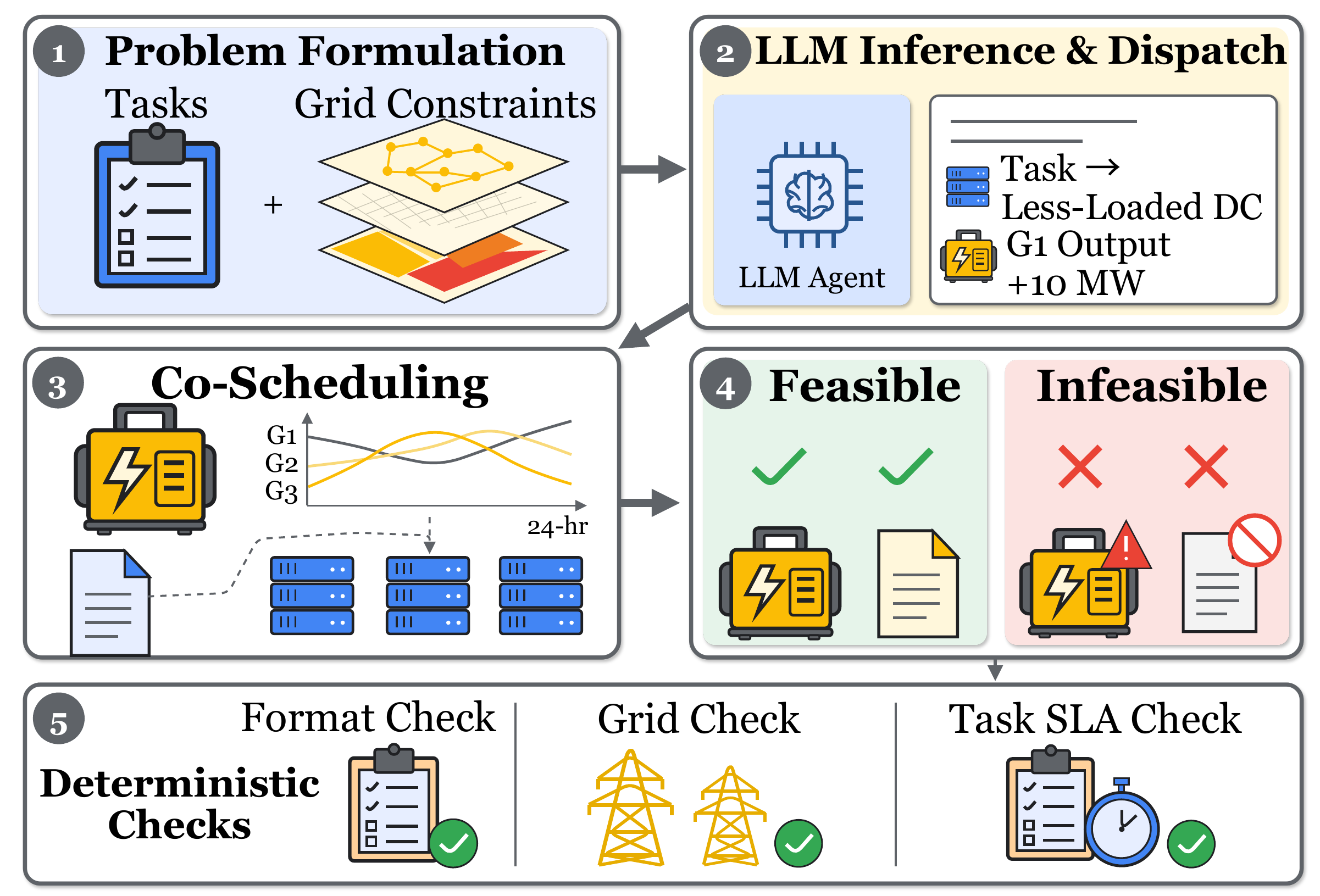}
  \caption{Illustrative ECCS workflow, from tasks and grid constraints to joint
  dispatch and placement. Deterministic format, grid, and task-service checks assess
  candidate schedules; actions and outcomes are schematic.}
  \label{fig:framework}
  \Description{Task requirements and grid constraints guide an LLM agent's joint power-dispatch and task-placement decisions. Feasible and infeasible outcomes are illustrated alongside deterministic format, grid, and task-service checks.}
\end{figure}

% Figure 1 is single-column; Figures 2--4 span both columns.
\begin{figure*}[t]
  \centering
  \includegraphics[width=\textwidth]{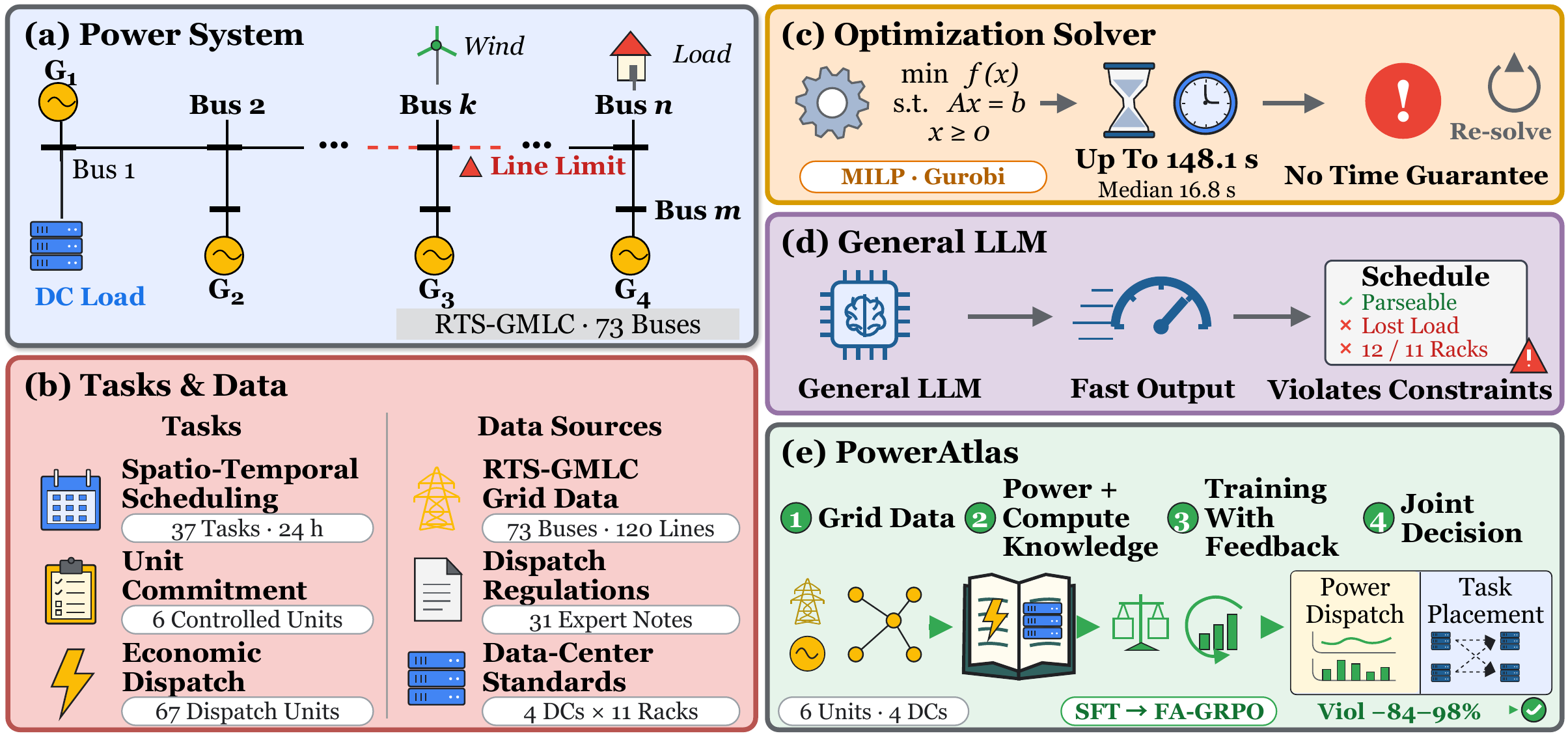}
  \caption{ECCS motivation and PowerAtlas overview. (a--b) Grid constraints and
  task/data sources define the problem. (c--d) Optimization latency and
  constraint-violating LLM outputs motivate (e) learning joint dispatch and
  placement from domain knowledge and training feedback.}
  \label{fig:motivation}
  \Description{The power-system and task-data panels define ECCS. The optimization-solver panel shows variable solve time, while the general-LLM panel illustrates fast outputs that violate constraints. PowerAtlas combines grid data, power and computing knowledge, and training feedback to generate joint decisions.}
\end{figure*}

\begin{figure*}[t]
  \centering
  \includegraphics[width=\textwidth]{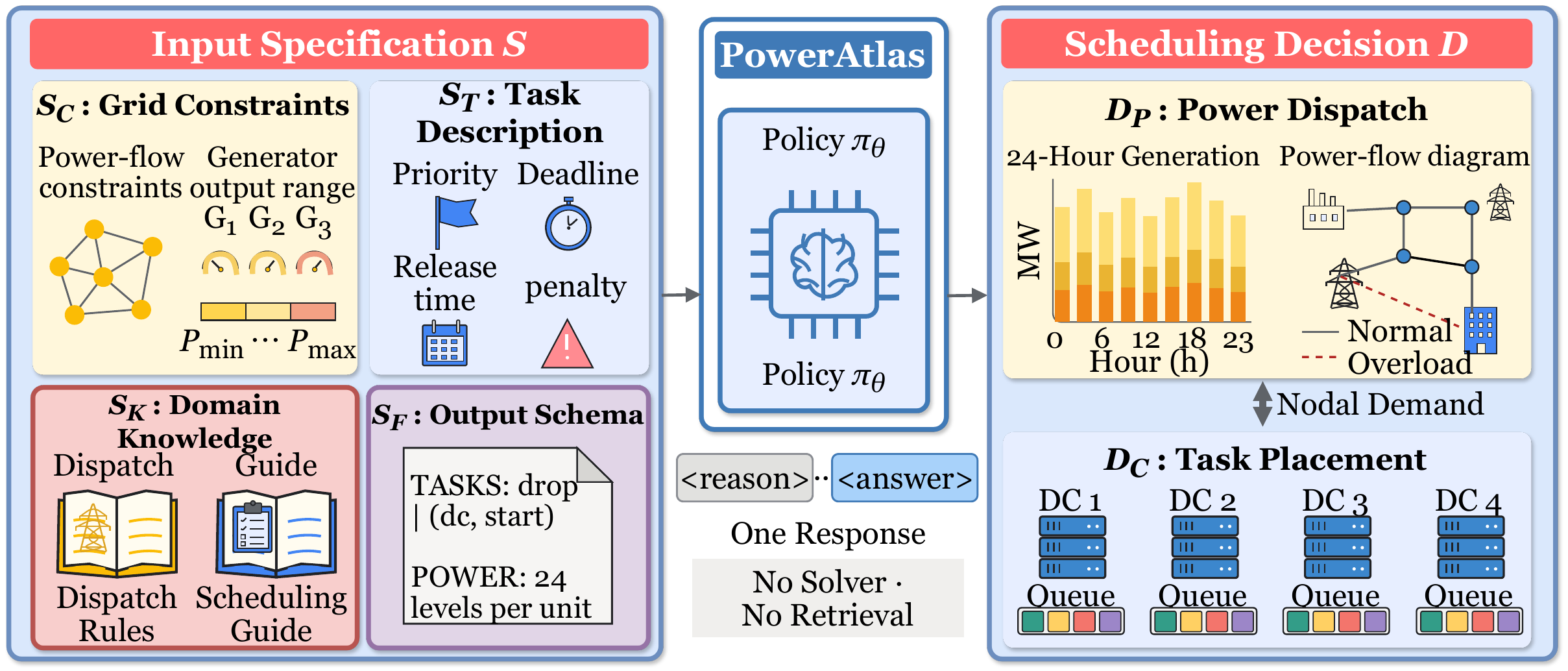}
  \caption{PowerAtlas decision interface. The policy maps
  $S=(S_C,S_T,S_K,S_F)$ to one response containing the joint decision
  $D=(D_P,D_C)$. Nodal demand couples 24-hour power dispatch with task placement
  across four data centers. Reported inference uses neither solver calls nor retrieval.}
  \label{fig:task}
  \Description{Grid constraints, task attributes, domain knowledge, and an output schema form the input specification. PowerAtlas emits reason and answer blocks in one response. Power dispatch and task placement are coupled through nodal demand; inference uses no solver or retrieval.}
\end{figure*}

\begin{figure*}[t]
  \centering
  \includegraphics[width=\textwidth]{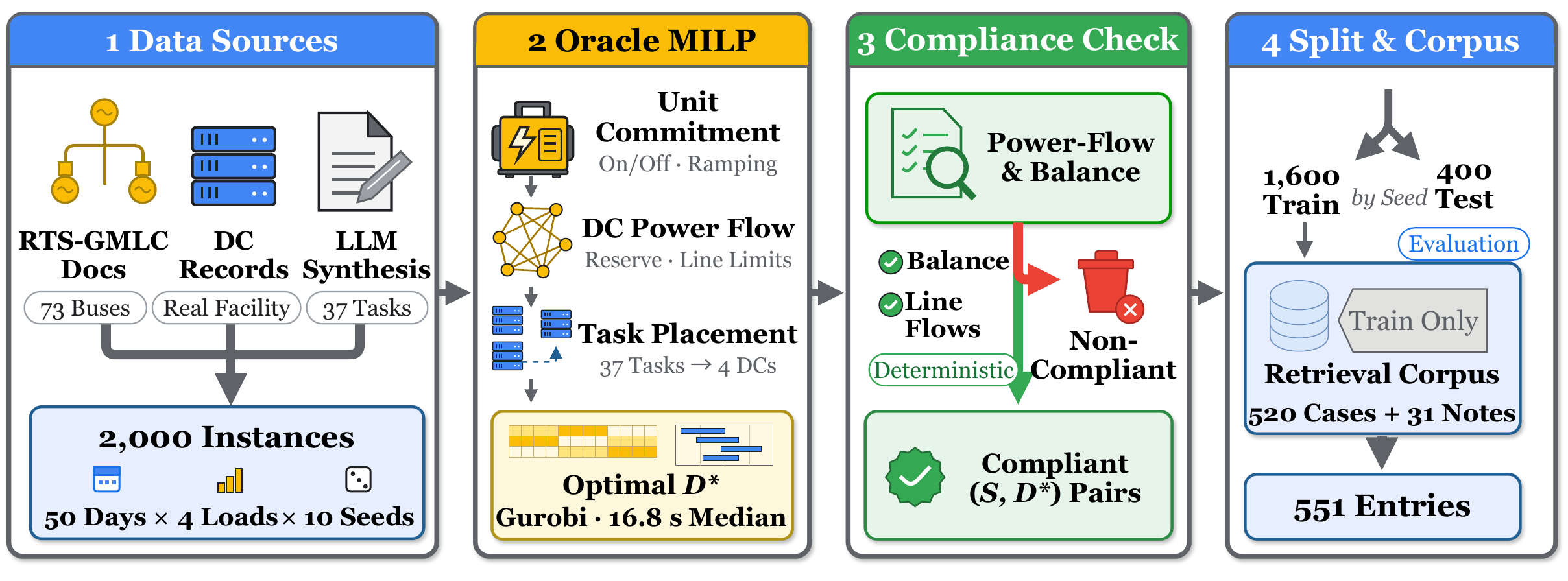}
  \caption{ECBench construction: data sources yield $2{,}000$ instances, a joint
  MILP supplies optimal schedules, and compliance checks validate instance--solution
  pairs. A seed-based $1{,}600/400$ train/test split supplies a training-only
  retrieval corpus of $520$ solved training cases and $31$ expert notes.}
  \label{fig:construction}
  \Description{RTS-GMLC documentation, data-center records, and LLM-assisted synthesis produce 2,000 instances from 50 days, four load levels, and ten seeds. A joint MILP yields reference schedules. Power-flow and balance checks precede a 1,600/400 training/test split; 520 training cases and 31 notes form the 551-entry retrieval corpus.}
\end{figure*}

Electricity--Computing Co-Scheduling (ECCS) jointly schedules computing tasks and
power dispatch: task placement changes nodal demand, while power availability
determines whether placements can be served (Figure~\ref{fig:framework}). Growing AI
workloads increase data-center electricity demand~\cite{IEA2025EnergyAI}.
Shifting jobs across times and sites~\cite{Lindberg2021Geographical,Lindberg2022GeoShift,Wiesner2021WaitAwhile,Williams2026PowerFlexible,Zhang2024Aggregator,Shahout2025Embedding}
offers flexibility, subject to available grid capacity and prices~\cite{Wang2023ComputingLoad,Chen2023Bidding}.

Prior co-scheduling research coordinates this flexibility through optimization
and learning~\cite{Zhang2025Mitigating,Hou2025Joint,Wan2026Data,Fan2026Harnessing,Liu2026Watts,Luo2025Novel},
building on unit commitment, dispatch, and power-flow models~\cite{Chen2023SCUC,Aharwar2023Unit,Wuijts2024Modelling,Fu2021Distributed,Kunya2023Economic,Xu2023Privacy,Zhao2022LineLosses,Gao2023DCPowerFlow,Taheri2024DCPowerFlow,Barrows2020RTS}.
More recently, LLMs~\cite{Yang2025Qwen3,Grattafiori2024Llama3,TII2024Falcon3}
have supported reasoning~\cite{Wei2022CoT,Yao2023ReAct}, optimization~\cite{Huang2025Orlm,Yang2024OPRO,AhmadiTeshnizi2024OptiMUS},
and power-system applications~\cite{Mohammadi2025Large,Bernier2025Powergraph,YangLarge,Jia2025Enhancing,Jadhav2025Evaluating,Zhou2024Elecbench}.

Two challenges remain (Figure~\ref{fig:motivation}). Coupled optimization can incur
variable solve times, whereas fast LLM outputs can be syntactically valid yet
violate grid or service constraints. Learning a response schema alone does not
resolve these dependencies. We therefore ask whether verifiable training feedback
can improve direct generation of jointly admissible schedules.

We formulate ECCS as a structured decision task and introduce \textbf{ECBench},
with $2{,}000$ calibrated instances, Gurobi-optimal reference schedules, and
executable metrics for grid constraints, service, cost, and latency.
Our method, \textbf{PowerAtlas}, combines supervised initialization with
\textbf{Feasibility-Aware Policy Alignment}. Domain knowledge and deterministic
scoring support training; inference produces one joint schedule without
retrieval, solver calls, or repair at decision time.

Across three open-weight backbones, paired comparisons on $10$--$40$ instances show
$89$--$98\%$ lower mean violation and up to $26\times$ lower penalized cost.
Archived stage comparisons characterize initialization and alignment in
output completeness and scheduling quality. We further analyze residual supply
deficits, checkpoint sensitivity, and service trade-offs; Section~\ref{sec:exp-setup}
specifies the sampling and satisfaction conventions for these comparisons.

\section{Related Work}

\noindent\textbf{Electricity-Computing Co-Scheduling.}
Workload shifting reduces costs~\cite{Radovanovi2023Carbon,Wiesner2021WaitAwhile,Liu2026Watts,Nkwawir2025Carbon}.
Prior formulations combine grid signals~\cite{Zhang2022GridSignals,Williams2026PowerFlexible}
with hierarchical, spatial, thermal, or multi-energy constraints~\cite{Han2024FourLevel,Su2025Cascade,Ali2024Spatial,Liu2024Collaborative,Bian2024EnergySharing}
and optimization or reinforcement learning~\cite{Luo2025Novel,Qi2022Microgrid}.
ECBench evaluates learned policies on coupled instances.

\noindent\textbf{LLM-Driven Scheduling and Optimization.}
LLM optimization~\cite{Abgaryan2024Llms,Xiao2024ChainExperts}
draws on retrieval~\cite{Luo2024ChatKBQA,Luo2025HyperGraphRAG,Luo2025KBQAO1,Luo2026GraphR1},
reasoning and tools~\cite{Wei2022CoT,Yao2023ReAct,Schick2023Toolformer,Yao2023TreeOfThoughts,Shinn2023Reflexion},
and reinforcement learning with verification~\cite{Ouyang2022InstructGPT,Schulman2017PPO,Lightman2024Verify,Guo2025DeepSeekR1}.
Power applications include scheduling~\cite{Mongaillard2024PowerScheduling},
dispatch~\cite{Mohammadi2025Large,Bernier2025Powergraph,YangLarge,Jia2025Enhancing,Cheng2025GAIA,Sarwar2025Large},
security~\cite{Ruan2024Security}, benchmarks~\cite{Zhou2024Elecbench,Han2022ADBench},
topology modeling~\cite{Chen2025Powerformer}, and physical verification~\cite{Jiang2026ScientificModeling}.
PowerAtlas targets direct joint scheduling with executable training feedback on the generated dispatch and placement.

\section{Task Definition}
\label{sec:task}

ECCS generates joint power dispatch and task placement to minimize operating cost
under grid and service constraints. Format validity and physical feasibility are
separate requirements (Figure~\ref{fig:task}).

The input $S=(S_C,S_T,S_K,S_F)$ specifies grid constraints, task requirements, domain
knowledge, and the output schema. Figure~\ref{fig:task} separates these components:
$S_C$ describes the grid, $S_T$ the task attributes, $S_K$ dispatch rules and scheduling
guidance, and $S_F$ the response contract. Grid constraints include DC power flow,
line and generator limits, ramping, and minimum up/down times. RTS-GMLC~\cite{Barrows2020RTS}
has 73 buses, 120 lines, and 73 thermal units: 6 are agent-controlled and 67 follow
economic dispatch. The 37 tasks specify priority, time windows, rack and power demand,
duration, drop penalties, and migratability across data-center locations.

The decision $D=(D_P,D_C)$ contains 24-hour schedules for five non-swing units, expressed
as \texttt{off} or percentages in $[0,100]$ mapped to $[P_{\min},P_{\max}]$.
The evaluator re-optimizes the swing unit within its capacity.
Each task is assigned \texttt{drop} or $(\mathrm{dc},\mathrm{start})$ across four
data centers, each with 11 racks and idle power $P_{\text{idle}}=0.25\,P_{\max}$.
Placement changes nodal demand and dispatch feasibility; Appendix~\ref{app:protocol}
specifies \mbox{incomplete-output handling.}

Let $C(D;S)$ include generation, lost-load, drop, and migration costs, and
$\mathcal{F}(S)$ collect the grid and service constraints. The oracle solves
\begin{equation}
D^{\star}=\arg\min_{D\in\mathcal{F}(S)}\ C(D;S).
\end{equation}
PowerAtlas approximates this optimization with a policy $\pi_{\theta}$ parameterized
by $\theta$,
\begin{equation}
D=\pi_{\theta}(S),
\end{equation}
The policy is trained to reduce violations and cost, but does not guarantee
$D\in\mathcal{F}(S)$. The hourly evaluator does not check ramping, minimum up/down
times, or reserve; its scores therefore do not certify full ECCS feasibility.
Section~\ref{sec:method} gives the training objective and schedule-level rewards.

\section{Benchmark: ECBench}
\label{sec:bench}

ECBench provides training instances, optimal reference schedules, and executable evaluation
of structured LLM decisions under coupled grid \mbox{and task-service constraints.}

\subsection{Dataset Construction}
\label{sec:bench-data}

Figure~\ref{fig:construction} summarizes ECBench construction. RTS-GMLC documentation,
data-center records, and LLM-assisted generation yield $2{,}000$ instances from
$50$ days, $4$ load levels, and $10$ seeds (Table~\ref{tab:ecbench-stats}).
Rack count and peak-load normalization are calibrated from a real facility;
task attributes are synthetic, with relaxed scheduling restrictions.

A joint unit-commitment and DC power-flow MILP models task placement, reserve,
data-center capacity, and nodal balance (Appendix~\ref{app:milp}).
Gurobi~\cite{Gurobi2024} solves it to optimality in $16.8$\,s at the median,
$35.9$\,s at the $95$th percentile, and up to $148.1$\,s.
Deterministic power-flow and balance checks validate instance--solution pairs.
The seed split contains $1{,}600$ training and $400$ test instances.
The retrieval corpus contains $520$ solved training cases and $31$ expert notes,
for $551$ entries; test cases are excluded (Appendix~\ref{app:corpus}).
The seed-based split shares operating regimes; Appendix~\ref{app:protocol}
specifies reporting-pool membership and checkpoint selection.

\subsection{Benchmark Design}
\label{sec:bench-design}

A deterministic evaluator scores each test instance with the agent's dispatch fixed.
For each hour $t$ of the horizon $T=24$, it solves a nodal DC power-flow feasibility
linear program (LP) over the bus set $\mathcal{N}$ and line set $\mathcal{L}$.
The LP minimizes the lost-load, over-generation, and line-overload slacks
$p^{\downarrow}_{n,t}$, $p^{\uparrow}_{n,t}$, and $f^{+}_{\ell,t}$.
These slacks and the submitted decision determine four groups of metrics.

The \emph{violation score} (Viol, in MWh-equivalent units; lower is better) combines
grid slacks and task-capacity excesses. Hourly slacks in MW are accumulated over the $24$-hour horizon,
\begin{equation}
\begin{aligned}
V_{\mathrm{grid}}&=\textstyle\sum_{n,t}(p^{\downarrow}_{n,t}+p^{\uparrow}_{n,t})
                  +\sum_{\ell,t}f^{+}_{\ell,t},\\
\mathrm{Viol}&=V_{\mathrm{grid}}+\mathrm{Viol}_{\mathrm{task}}.
\end{aligned}
\label{eq:viol}
\end{equation}
where $V_{\mathrm{grid}}$ sums the three grid slacks over all buses or lines and
$t=1,\ldots,T$, and $\mathrm{Viol}_{\mathrm{task}}$ is the MWh-equivalent task violation.
Viol is averaged over evaluated instances.
\emph{Computing satisfaction} credits tasks $\mathcal{P}$ under two conventions.
Placement-only scores count retained assignments. Grid-aware scores additionally
exclude task-specific violations and unpowered hours; neither convention certifies
all capacity constraints (Appendix~\ref{app:coverage}). For $n$ tasks with value $v_i$,
\mbox{rack demand $r_i$, and duration $d_i$:}
\begin{equation}
\begin{gathered}
\mathrm{S\text{-}cnt}=100\,\tfrac{|\mathcal{P}|}{n},\quad
\mathrm{S\text{-}val}=100\,\tfrac{\sum_{i\in\mathcal{P}}v_i}{\sum_i v_i},\\[-1pt]
\mathrm{S\text{-}rh}=100\,\tfrac{\sum_{i\in\mathcal{P}}r_i d_i}{\sum_i r_i d_i}.
\end{gathered}
\end{equation}
The three rates measure task count, value, and rack-hours. S-H/M/L give the rates
for each priority tier. \emph{Cost} combines generation, grid-slack penalties,
task dropping, and migration:
\begin{equation}
\mathrm{Cost}=C_{\mathrm{gen}}+C_{\mathrm{mig}}
+\textstyle\sum_{i\in\mathcal{D}}v_i+C_{\mathrm{LOL}}V_{\mathrm{grid}},
\label{eq:cost}
\end{equation}
where $C_{\mathrm{gen}}$ is generation cost, $C_{\mathrm{LOL}}$ is the common grid-slack
penalty rate, $\mathcal{D}$ is the submitted drop set, and $C_{\mathrm{mig}}$ is migration
cost. The evaluator prices all three grid slacks at this rate; the oracle objective
prices lost load (Appendix~\ref{app:milp}). Re-scored cost fixes minimum-slack ties by minimizing swing output (Appendix~\ref{app:reaudit}).
\emph{Latency} reports mean wall-clock decision time under each model's serving
conditions, without dispersion estimates. Appendix~\ref{app:coverage} details the \mbox{evaluator's constraint coverage.}

\section{Methodology: PowerAtlas}
\label{sec:method}

Algorithm~\ref{alg:fagrpo} summarizes policy training using the reference schedules
constructed in Figure~\ref{fig:construction}. PowerAtlas learns the output schema and
scheduling patterns through supervised initialization, then refines the policy with
feasibility-aware policy alignment using domain-specific rewards. The available training scorer checks tasks and aggregate
power balance; final evaluation uses a nodal DC LP. These procedures have different
constraint coverage (Appendix~\ref{app:coverage}). Neither feeds corrections back
to the model at inference during schedule generation.

\subsection{Structured Decision Generation}
\label{sec:method-init}

For each instance, the policy $\pi_{\theta}$ maps the input specification $S$ to a joint
decision through the interface in Figure~\ref{fig:task}. The structured prompt combines
hourly grid signals (load, renewable generation, net load, and aggregate fixed
generation), task attributes, domain guidance, and the output contract.
The decision $D=(D_P,D_C)$ is emitted in one response. The policy $\pi_{\theta}$
decodes it autoregressively as a templated token sequence $y$,
\begin{equation}
\pi_{\theta}(D\mid S)=\prod_{k=1}^{|y|}\pi_{\theta}\!\big(y_k\mid y_{<k},\,S\big),
\label{eq:policy}
\end{equation}
The response contains a \texttt{<reason>} rationale~\cite{Wei2022CoT} and an
\texttt{<answer>} block: \texttt{TASKS} assigns \texttt{drop} or
$(\mathrm{dc},\mathrm{start})$; \texttt{POWER} gives 24 values per adjustable unit.
Training may retrieve cases through Graph-R1~\cite{Luo2026GraphR1,Luo2025HyperGraphRAG};
reported \mbox{inference uses no retrieval.}

\subsection{Feasibility-Aware Policy Alignment}
\label{sec:method-grpo}

Full-parameter supervised fine-tuning on solved instances $(S,D^{\star})$
minimizes response-token cross-entropy for the target $y^{\star}$,
\begin{equation}
\mathcal{L}_{\mathrm{SFT}}(\theta)=-\,\mathbb{E}_{(S,D^{\star})}\sum_{k=1}^{|y^{\star}|}\log\pi_{\theta}\!\big(y^{\star}_k\mid y^{\star}_{<k},\,S\big),
\label{eq:sft}
\end{equation}
Only response tokens contribute to the loss; the specification provides context.

Feasibility-aware policy alignment assigns a terminal reward $R(\tau)$ to the completed response
$\tau=y_{1:|y|}$ and maximizes $\mathbb{E}_{\tau\sim\pi_{\theta}}[R(\tau)]$.
The token-level state is $(S,y_{<k})$ and the action is the next token $y_k$.
For each instance $q$, the behavior policy $\pi_{\theta_{\mathrm{old}}}$ samples
$N$ trajectories $\{\tau_i\}_{i=1}^{N}$. We optimize a critic-free, group-relative
objective in the style of GRPO~\cite{Shao2024DeepSeekMath}, replacing the value network
of PPO~\cite{Schulman2017PPO} with a group baseline. Following
Graph-R1~\cite{Luo2026GraphR1} and reinforcement-learning alignment of reasoning
LLMs~\cite{Ouyang2022InstructGPT,Guo2025DeepSeekR1}, we use the objective
\begin{equation}
\label{eq:grpo}
\begin{aligned}
\mathcal{J}(\theta)=\mathbb{E}\Bigg[&\frac{1}{N}\sum_{i=1}^{N}\frac{1}{|\tau_i|}\sum_t\\ &\min\!\big(\rho_{\theta}\,\hat{A}_i,\ \mathrm{clip}(\rho_{\theta},1\pm\epsilon)\,\hat{A}_i\big)\\
&-\,\beta\,\mathbb{D}_{\mathrm{KL}}(\pi_{\theta}\|\pi_{\mathrm{ref}})\Bigg],
\end{aligned}
\end{equation}

\begin{algorithm}[!t]
\footnotesize
\caption{Feasibility-Aware Policy Alignment}
\label{alg:fagrpo}
\begin{algorithmic}[1]
\Require solved instances $\mathcal{D}$; group size $N$; KL weight $\beta$; reward weights $(w_c,w_p,\eta_v,\eta_c)$; training penalty $\lambda_{\mathrm{tr}}$; format penalty $r_0$.
\Ensure Aligned policy $\pi_{\theta}(\cdot\mid S)$.
\State \textbf{Stage I (SFT):} initialize $\theta$ from a pretrained model;
\While{\textit{not converged}}
  \State sample minibatch $\mathcal{B}\subset\mathcal{D}$;
  \State update $\theta\leftarrow\theta-\eta\nabla_{\theta}\mathcal{L}_{\mathrm{SFT}}(\theta;\mathcal{B})$ using Eq.~\eqref{eq:sft};
\EndWhile
\State freeze $\pi_{\mathrm{ref}}\leftarrow\pi_{\theta}$;
\Statex
\State \textbf{Stage II (Policy alignment):}
\While{\textit{not converged}}
  \State sample a batch of instances $\mathcal{Q}=\{q\}$;\ set $\theta_{\mathrm{old}}\leftarrow\theta$;
  \For{each $q\in\mathcal{Q}$}
    \State sample $N$ trajectories $\{\tau_i\}_{i=1}^{N}\sim\pi_{\theta_{\mathrm{old}}}(\cdot\mid S)$;
    \For{each $\tau_i$}
      \State parse $D_i$; if ill-formed, $R(\tau_i)\!\leftarrow\!-r_0$ (format gate);
      \State else apply training-side task and power scoring: $V_{\mathrm{tr},i},\kappa_i,\phi_i,c_{\theta,i}^{\mathrm{tr}}$;
      \State compute $Q_{\mathrm{compute}}$ via Eq.~\eqref{eq:qcompute} and $Q_{\mathrm{grid}}$ via Eq.~\eqref{eq:qgrid};
      \State aggregate $R(\tau_i)$ via Eqs.~\eqref{eq:rquality} and~\eqref{eq:rgate};
    \EndFor
    \State compute $\bar{R},s_R$ and advantages $\hat{A}_i$ via Eq.~\eqref{eq:adv};
    \State update $\theta\leftarrow\arg\max_{\theta}\mathcal{J}(\theta)$ using Eq.~\eqref{eq:grpo};
  \EndFor
\EndWhile
\end{algorithmic}
\end{algorithm}
where $\epsilon$ is the clipping range and $\beta$ weights KL divergence.
The importance ratio and group-relative advantage are
\begin{equation}
\rho_{\theta}=\frac{\pi_{\theta}(\mathbf{a}_t^{(i)}\mid \mathbf{s}_{t-1}^{(i)})}{\pi_{\theta_{\mathrm{old}}}(\mathbf{a}_t^{(i)}\mid \mathbf{s}_{t-1}^{(i)})},
\end{equation}
\begin{equation}
\hat{A}_i=\frac{R(\tau_i)-\operatorname{mean}_{j}R(\tau_j)}{\operatorname{std}_{j}R(\tau_j)+\delta},
\label{eq:adv}
\end{equation}
where $\delta=10^{-6}$ stabilizes the denominator in the available implementation.
For a group of identical rewards, every advantage is zero.

\begin{table*}[t]

\centering
\caption{Baseline results: one instance for frontier models and 10--40 for open-weight models.
Satisfaction is uniformly grid-aware. Cost and latency retain archived evaluation settings; these rows are descriptive, not a paired ranking.}
\label{tab:main-alt}
\fontsize{9}{10}\selectfont
\setlength{\tabcolsep}{2pt}
\renewcommand{\arraystretch}{0.95}
\begin{tabularx}{\textwidth}{@{}>{\hsize=2.2\hsize\linewidth=\hsize\centering\arraybackslash}X>{\hsize=1.1\hsize\linewidth=\hsize\centering\arraybackslash}X>{\hsize=0.75\hsize\linewidth=\hsize\centering\arraybackslash}X>{\hsize=0.75\hsize\linewidth=\hsize\centering\arraybackslash}X>{\hsize=0.75\hsize\linewidth=\hsize\centering\arraybackslash}X>{\hsize=0.75\hsize\linewidth=\hsize\centering\arraybackslash}X>{\hsize=0.75\hsize\linewidth=\hsize\centering\arraybackslash}X>{\hsize=0.75\hsize\linewidth=\hsize\centering\arraybackslash}X>{\hsize=1.4\hsize\linewidth=\hsize\centering\arraybackslash}X>{\hsize=0.8\hsize\linewidth=\hsize\centering\arraybackslash}X@{}}
\toprule
\multirow{3}{*}{\textbf{Method}} & \textbf{Grid}
 & \multicolumn{6}{c}{\textbf{Computing satisfaction (\%)\,$\uparrow$}}
 & \textbf{Economy} & \textbf{Time}\\
\cmidrule(lr){2-2}\cmidrule(lr){3-8}\cmidrule(lr){9-9}\cmidrule(lr){10-10}
 & \textbf{Viol} & \multicolumn{3}{c}{\textbf{Overall}}
 & \multicolumn{3}{c}{\textbf{Tiered}} & \textbf{Cost} & \textbf{Latency}\\
\cmidrule(lr){3-5}\cmidrule(lr){6-8}
 & \textbf{(MWh)}\,$\downarrow$ & \textbf{S-cnt} & \textbf{S-val} & \textbf{S-rh}
 & \textbf{S-H} & \textbf{S-M} & \textbf{S-L}
 & \textbf{(\$)}\,$\downarrow$ & \textbf{(s)}\,$\downarrow$\\
\midrule
\textbf{DeepSeek-V4-Flash}
 & 5{,}234.2 & 20.3 & 23.4 & 19.0 & 26.5 & 20.0 & 10.0 & 47{,}948{,}178 & \textbf{12.6} \\
\textbf{DeepSeek-V4-Pro}
 & 2{,}763.4 & 20.3 & 22.1 & 18.1 & 29.4 & 20.0 & 5.0 & 25{,}762{,}501 & 17.3 \\
\textbf{Kimi-K2.6 (fast)}
 & 1{,}212.0 & 50.0 & 55.5 & 51.0 & 58.8 & 60.0 & 25.0 & 12{,}392{,}536 & 28.3 \\
\textbf{Kimi-K2.6 (pro)}
 & 1{,}488.6 & 54.1 & 53.4 & 50.2 & 58.8 & 60.0 & 40.0 & 14{,}838{,}689 & 34.4 \\
\textbf{Qwen3.7-Plus}
 & 741.0 & 79.7 & 83.1 & 80.5 & \textbf{88.2} & 65.0 & \textbf{80.0} & 9{,}344{,}901 & 31.0 \\
\textbf{Qwen3.7-Max}
 & 462.6 & \textbf{82.0} & \textbf{86.2} & \textbf{83.1} & 86.3 & \textbf{96.7} & 60.0 & \textbf{5{,}725{,}763} & 29.9 \\
\hdashline
\textbf{GPT-5.5}
 & \textbf{445.8} & 64.3 & 66.9 & 63.5 & 70.6 & 64.0 & 54.0 & 6{,}068{,}833 & 14.1 \\
\textbf{Claude Haiku 4.5}
 & 543.3 & 67.6 & 57.6 & 65.7 & 56.9 & 74.8 & 78.6 & 7{,}548{,}683 & 13.9 \\
\textbf{Qwen3-4B-Instruct}
 & 1{,}766.9 & 55.7 & 50.9 & 52.1 & 52.5 & 61.5 & 56.0 & 17{,}706{,}710 & 67.4 \\
\textbf{Llama-3.2-3B-Instruct}
 & 6{,}492.1 & 12.2 & 11.8 & 10.6 & 13.6 & 15.4 & 8.1 & 59{,}997{,}272 & 30.1 \\
\textbf{Falcon3-3B-Instruct}
 & 11{,}079.2 & 3.0 & 3.0 & 2.8 & 3.1 & 4.0 & 2.1 & 101{,}484{,}114 & 19.6 \\
\bottomrule
\end{tabularx}

\par\vspace{8pt}

\centering
\caption{Paired quality and cost results: Qwen $n=10$, Llama $n=12$, Falcon $n=40$; grid-aware satisfaction and the fixed LP tie-break (Appendix~\ref{app:reaudit}). Latency retains original-run means: base $n=10/12/40$, PowerAtlas $n=40$.}
\label{tab:backbone}
\fontsize{9}{10}\selectfont
\setlength{\tabcolsep}{2pt}
\renewcommand{\arraystretch}{0.95}
\begin{tabularx}{\textwidth}{@{}>{\hsize=2.2\hsize\linewidth=\hsize\centering\arraybackslash}X>{\hsize=1.1\hsize\linewidth=\hsize\centering\arraybackslash}X>{\hsize=0.75\hsize\linewidth=\hsize\centering\arraybackslash}X>{\hsize=0.75\hsize\linewidth=\hsize\centering\arraybackslash}X>{\hsize=0.75\hsize\linewidth=\hsize\centering\arraybackslash}X>{\hsize=0.75\hsize\linewidth=\hsize\centering\arraybackslash}X>{\hsize=0.75\hsize\linewidth=\hsize\centering\arraybackslash}X>{\hsize=0.75\hsize\linewidth=\hsize\centering\arraybackslash}X>{\hsize=1.4\hsize\linewidth=\hsize\centering\arraybackslash}X>{\hsize=0.8\hsize\linewidth=\hsize\centering\arraybackslash}X@{}}
\toprule
\multirow{3}{*}{\textbf{Method}} & \textbf{Grid}
 & \multicolumn{6}{c}{\textbf{Computing satisfaction (\%)\,$\uparrow$}}
 & \textbf{Economy} & \textbf{Time}\\
\cmidrule(lr){2-2}\cmidrule(lr){3-8}\cmidrule(lr){9-9}\cmidrule(lr){10-10}
 & \textbf{Viol} & \multicolumn{3}{c}{\textbf{Overall}}
 & \multicolumn{3}{c}{\textbf{Tiered}} & \textbf{Cost} & \textbf{Latency}\\
\cmidrule(lr){3-5}\cmidrule(lr){6-8}
 & \textbf{(MWh)}\,$\downarrow$ & \textbf{S-cnt} & \textbf{S-val} & \textbf{S-rh}
 & \textbf{S-H} & \textbf{S-M} & \textbf{S-L}
 & \textbf{(\$)}\,$\downarrow$ & \textbf{(s)}\,$\downarrow$\\
\midrule
\textbf{Qwen3-4B-Instruct} & 1{,}766.9 & 55.7 & 50.9 & 52.1 & 52.5 & 61.5 & 56.0 & 17{,}615{,}960 & 67.4 \\
\rowcolor{resultlavender}
\textbf{\quad + PowerAtlas} & \textbf{188.5} & \textbf{59.5} & \textbf{79.0} & \textbf{56.1} & \textbf{91.8} & \textbf{34.5} & \textbf{3.1} & \textbf{3{,}648{,}432} & \textbf{57.2} \\
\midrule
\textbf{Llama-3.2-3B-Instruct} & 6{,}492.1 & 12.2 & 11.8 & 10.6 & 13.6 & 15.4 & 8.1 & 59{,}957{,}692 & 30.1 \\
\rowcolor{resultlavender}
\textbf{\quad + PowerAtlas} & \textbf{213.7} & \textbf{78.4} & \textbf{76.4} & \textbf{72.6} & \textbf{81.9} & \textbf{73.3} & \textbf{68.3} & \textbf{2{,}992{,}431} & \textbf{18.5} \\
\midrule
\textbf{Falcon3-3B-Instruct} & 11{,}079.2 & 3.0 & 3.0 & 2.8 & 3.1 & 4.0 & 2.1 & 101{,}469{,}344 & 19.6 \\
\rowcolor{resultlavender}
\textbf{\quad + PowerAtlas} & \textbf{216.6} & \textbf{69.5} & \textbf{65.2} & \textbf{62.9} & \textbf{71.7} & \textbf{59.6} & \textbf{66.9} & \textbf{3{,}830{,}536} & \textbf{18.8} \\
\bottomrule
\end{tabularx}
\end{table*}

Algorithm~\ref{alg:fagrpo} summarizes the base reward and update. Training-side
task and power scores give
\begin{equation}
\label{eq:rquality}
R_{\mathrm{quality}}=w_c\,Q_{\mathrm{compute}}+w_p\,Q_{\mathrm{grid}},
\end{equation}
weighted by $w_c,w_p$. Task quality discounts retained value and count when task
checks fail:
\begin{equation}
\label{eq:qcompute}
Q_{\mathrm{compute}}=\kappa\,\phi\,\big(\eta_v\,s_{\mathrm{val}}+\eta_c\,s_{\mathrm{cnt}}\big),
\end{equation}
where $s_{\mathrm{val}},s_{\mathrm{cnt}}\in[0,1]$ are retained value and count fractions
from the training scorer, and $\kappa$ is the fraction of tasks with a decision.
The task-check discount is $\phi=1$ when those checks pass and $\phi=\phi_0<1$
otherwise. The separate grid-quality term is
\begin{equation}
\label{eq:qgrid}
Q_{\mathrm{grid}}=\min\!\Big(1,\ \frac{c^{\star}_{\mathrm{tr}}}{c^{\mathrm{tr}}_{\theta}+\lambda_{\mathrm{tr}}V_{\mathrm{tr}}}\Big),
\end{equation}
where the available fast scorer uses generation costs of the six controlled units
for $c^{\star}_{\mathrm{tr}}$ and $c^{\mathrm{tr}}_{\theta}$. It penalizes the aggregate
balance residual $V_{\mathrm{tr}}$ at rate $\lambda_{\mathrm{tr}}$, separately from
the final LP metrics. A format gate gives the base reward
\begin{equation}
\label{eq:rgate}
R(\tau)=
\begin{cases}
R_{\mathrm{quality}}, & \text{if the output is well-formed,}\\[2pt]
-\,r_0, & \text{otherwise,}
\end{cases}
\end{equation}
% Figure 5 -- results figure (a/b top, c-f bottom).
% ============================================================================
% Figure 5 -- results (2 wide on top + 4 narrow below).
% Requires in the preamble:  \usepackage{graphicx}  \usepackage{tikz}
%
% top     (a) fig5_1  performance landscape   (b) fig5_cd single-instance traces
% bottom  (c) fig5_traj training dynamics     (d) fig5_tierstage tiers over training
%         (e) fig5_robust robustness          (f) fig5_paired per-instance before/after
%
% Every panel is drawn 1:1 at the width of its slot by plot_fig5_panels.py, so
% the type size in the PDF is the type size on the page (~6.5-7 pt); do not
% rescale the \includegraphics widths without regenerating the panels.
% The panel letter is a zero-height overlay pinned to the top-left corner of the
% image, so it costs no vertical space. Dashed rules separate (a) from (b) and
% the top row from the bottom row.
% ============================================================================
\newsavebox{\fivebox}
\newsavebox{\fivetopbox}
% The letter is drawn as a TikZ node *on top of* the image (the panels have an
% opaque white background, so anything placed before the image is hidden by it).
\newcommand{\panelfig}[3]{% #1 slot width, #2 letter, #3 file
  \begin{minipage}[t]{#1}\vspace{0pt}%
    \begin{tikzpicture}[inner sep=0pt, outer sep=0pt]
      \node[anchor=north west] (img) at (0,0)
        {\includegraphics[width=\linewidth]{#3}};
      \node[anchor=north west, inner sep=0pt, font=\fontsize{8.3}{9}\selectfont\bfseries,
            yshift=2.2pt] at (img.north west) {(#2)};
    \end{tikzpicture}%
  \end{minipage}}

\begin{figure*}[t]
\centering
% measured once so the vertical rule is exactly as tall as the panels beside it
\sbox{\fivetopbox}{\includegraphics[width=0.495\textwidth]{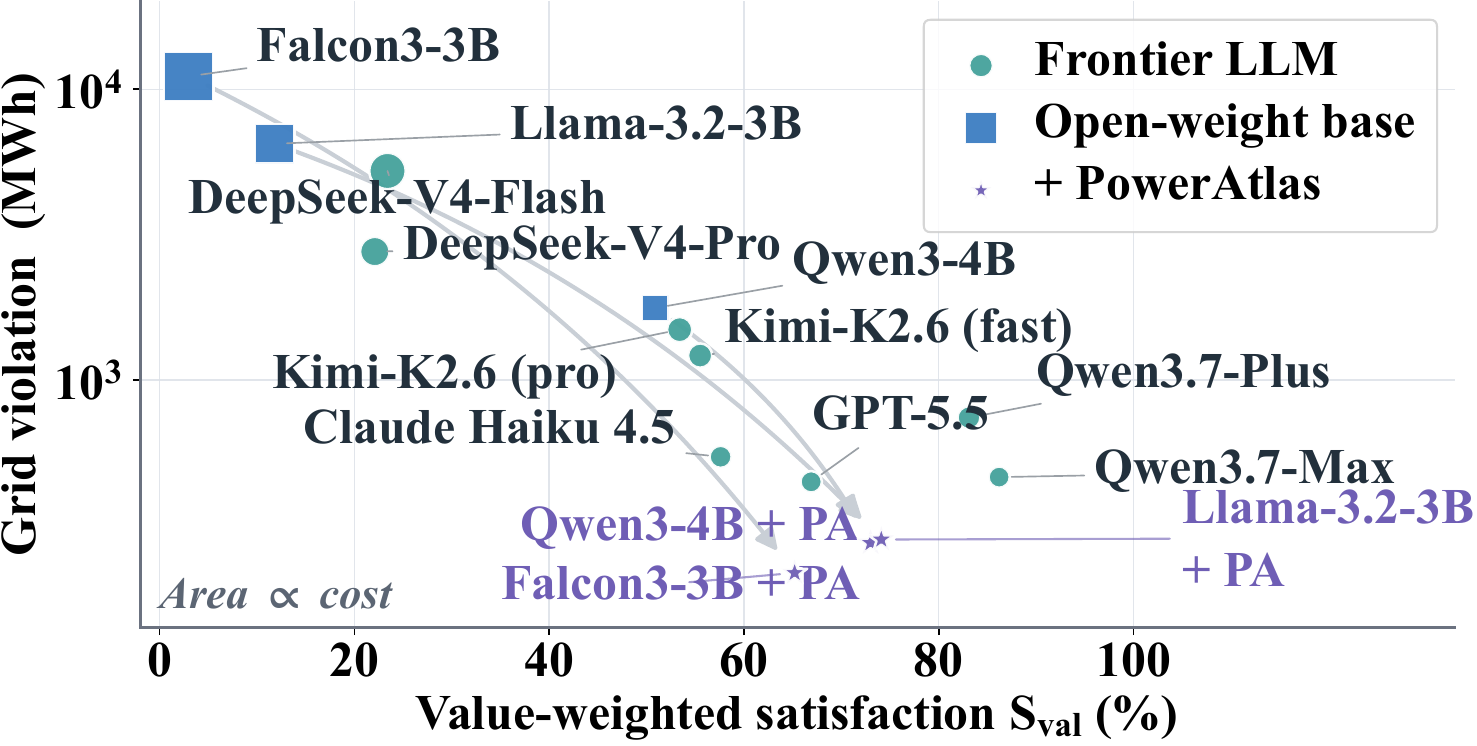}}%

% ------------------------------------------------------------------ top row
\panelfig{0.495\textwidth}{a}{figure/panels/fig5_1.pdf}\hfill
\begin{minipage}[t]{0.014\textwidth}\vspace{0pt}%
  \centering
  % overlay: the rule is drawn past the ends of its own box, so lengthening it cannot grow the row
  % with overlay the origin sits at the top of the (now zero-size) box, so the rule runs downwards
  \tikz[overlay]{\draw[densely dashed, line width=0.5pt, black!38]
        (0,16pt) -- (0,-\dimexpr\ht\fivetopbox+12pt\relax);}%
\end{minipage}\hfill
\panelfig{0.4638\textwidth}{b}{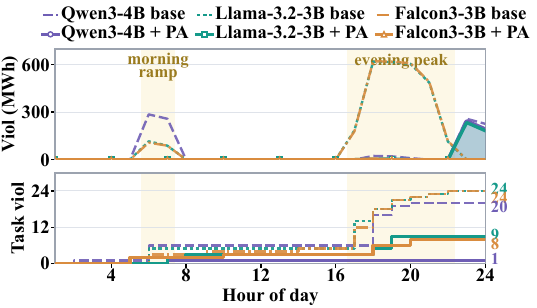}

\par\vspace{4pt}
\noindent\tikz{\draw[densely dashed, line width=0.5pt, black!38]
      (0,0) -- (\the\textwidth,0);}\par\nointerlineskip
\vspace{2pt}

% --------------------------------------------------------------- bottom row
\panelfig{0.2436\textwidth}{c}{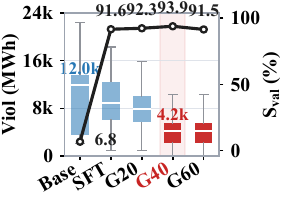}\hfill
\panelfig{0.1964\textwidth}{d}{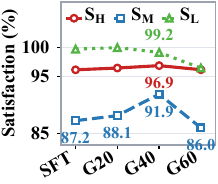}\hfill
\panelfig{0.2482\textwidth}{e}{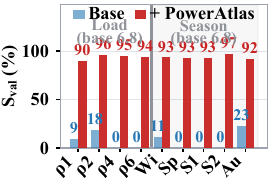}\hfill
\panelfig{0.2095\textwidth}{f}{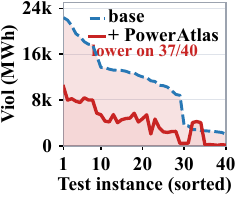}

\caption{Results on ECBench (PA denotes PowerAtlas). (a)~Violation versus grid-aware satisfaction (bubble size indicates archived cost; sample sizes differ).
  (b)~Hourly violation and cumulative task violations on one test instance, one line per system.
  (c)~Training dynamics, (d)~tier satisfaction across stages, (e)~$S_{val}$ by data-center
  load level and season, and (f)~base versus selected checkpoint for each instance. Panels
  (c)--(f) use a separate archived run and placement-based satisfaction; residuals in (c)/(f) are training-side proxies. Paired final-evaluator gains are in Table~2.}
\label{fig:fig5}
  \Description{Six result panels: violation versus satisfaction, hourly violation traces on one instance, training dynamics, tier satisfaction across checkpoints, satisfaction by data-center load level and season, and a per-instance before-and-after comparison.}
\end{figure*}

where $r_0>0$ is fixed. The available code also adds a capped retrieval bonus to
well-formed outputs; Appendix~\ref{app:impl} records this implementation detail and
the unresolved historical configuration mapping. Scoring does not guide reported inference or revise the generated joint schedule.

\section{Experiments}
\label{sec:exp}

We evaluate PowerAtlas on ECBench through four research questions
(Tables~\ref{tab:main-alt}--\ref{tab:stage}, Figures~\ref{fig:fig5} and~\ref{fig:cases}):
\textbf{RQ1}, can verifier-guided training improve constraint satisfaction in generated schedules;
\textbf{RQ2}, what do supervised initialization and feasibility-aware alignment each contribute;
\textbf{RQ3}, which constraint violations persist after training;
and \textbf{RQ4}, how do performance and service trade-offs vary across checkpoints and operating conditions?

\subsection{Experimental Setup}
\label{sec:exp-setup}

ECBench contains $2{,}000$ instances with Gurobi-optimal reference schedules.
The retrieval corpus uses only the training split. Eight frontier LLMs
(Table~\ref{tab:main-alt}) use the same prompt through an OpenAI-compatible API;
three untrained open-weight backbones provide baselines before domain training.
Supervised initialization is followed by up to $60$ alignment steps.
Reported inference uses greedy decoding in one response, without retrieval or a solver.
Main tables use grid-aware satisfaction; placement-only training diagnostics are
separately identified in Figure~\ref{fig:fig5} and Appendix~\ref{app:protocol}.
Missing units or hours are treated as offline. Latency includes the serving path:
remote APIs for frontier models and local vLLM for open-weight models.
Frontier results cover repeated runs on one shared high-load instance. Open-weight
baselines cover $10$, $12$, or $40$ instances; trained checkpoints use a $40$-instance
pool also used for selection. Table~\ref{tab:backbone} pairs identical IDs within
each backbone; these retrospective comparisons are not blind test results.

\subsection{Constraint Satisfaction in Generated Decisions (RQ1)}
\label{sec:exp-main}

\textbf{Constraint satisfaction.} Archived frontier outputs have mean violations of
$445.8$--$5{,}234.2$\,MWh on one instance. Table~\ref{tab:backbone} pairs base and
trained outputs on identical instances: Qwen ($n=10$), Llama ($n=12$), and Falcon
($n=40$) reach $188.5$, $213.7$, and $216.6$\,MWh, respectively.
Mean violations fall by $89.3\%$, $96.7\%$, and $98.0\%$; penalized cost falls by
$4.8\times$, $20.0\times$, and $26.5\times$ under a fixed evaluator.
All $10/10$, $12/12$, and $40/40$ pairs improve on final violation.

\textbf{Cost and runtime.} These gains compare learned policies with their own
base models. Supplementary fixed-background HiGHS controls pass hourly grid and
task checks on up to $40/40$ cases within a measured 20-second budget
(Appendix~\ref{app:reaudit}); optimization superiority is not established.
Historical LLM latency uses different serving systems.

\subsection{Contributions of Initialization and Alignment (RQ2)}
\label{sec:exp-abl}

\textbf{Supervised initialization.} Table~\ref{tab:stage} separates two archived
SFT outputs: the earlier archive has $9{,}084.2$\,MWh mean violation, while SFT-16
has $682.4$\,MWh and $43.6\%$ grid-aware value satisfaction.
Their initialization histories must not be conflated.

\textbf{Feasibility-aware alignment.} Relative to the SFT-16 archive, PowerAtlas
has $68.3\%$ lower mean violation, about $2.1\times$ lower cost, and $65.2\%$
grid-aware value satisfaction on identical instances. The no-initialization archive
remains at $10{,}730.2$\,MWh. These stage comparisons support the training sequence;
budget-matched reward and retrieval controls are still needed for causal attribution.

\begin{figure}[!t]
\centering
\includegraphics[width=\columnwidth]{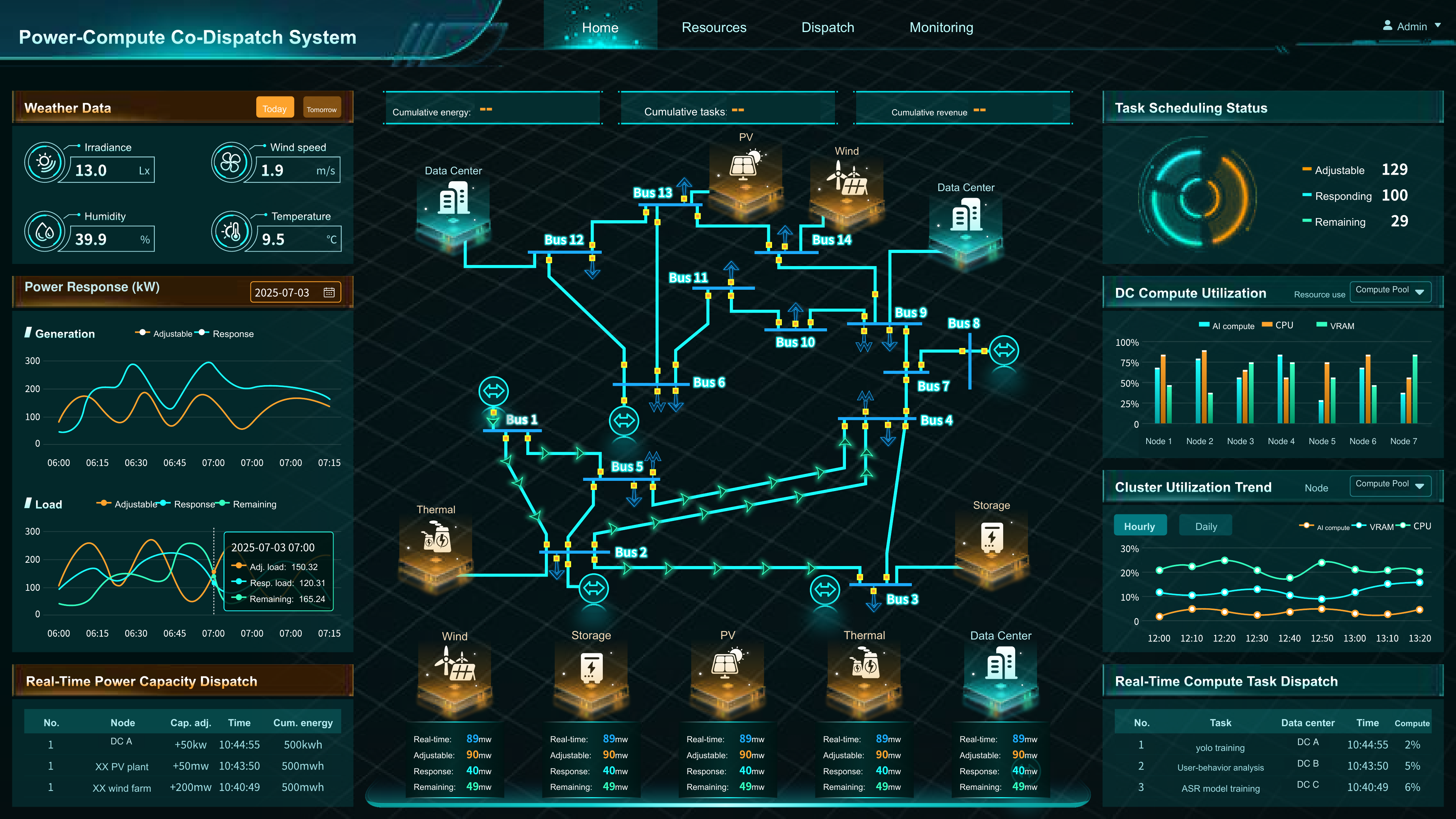}
\caption{Co-scheduling system interface showing grid state, computing-resource
utilization, and power and task dispatch. Identifying labels are replaced with
generic site and resource names.}
\label{fig:system}
\Description{A co-scheduling dashboard showing power-network connections, resource utilization, and power and computing-task dispatch views, with generic site and resource labels.}
\end{figure}

% ============================================================================
% Figure 6 -- failure-case study (2x2 cards, full width).
% Needs: xcolor, tcolorbox[skins], tikz, and the Comic Sans MS setup in the
% preamble (\comicsans). All excerpts are verbatim from the stored model
% answers in pred_*/ ; every violation quoted here is the one the evaluator
% reports.
% ============================================================================

% ---- per-case palettes (frame / title bar / analysis tint) -----------------
\definecolor{caseA}{HTML}{D33F4E}   % Case 1 -- crimson
\definecolor{caseB}{HTML}{E08A1E}   % Case 2 -- amber
\definecolor{caseC}{HTML}{7A54C4}   % Case 3 -- violet
\definecolor{caseD}{HTML}{169C93}   % Case 4 -- teal
% ---- accent colours for the diagnosis pill (one per case) ------------------
\definecolor{pillA}{HTML}{3B3F8F}   % indigo
\definecolor{pillB}{HTML}{1C6E8C}   % petrol blue
\definecolor{pillC}{HTML}{2E7D4F}   % forest green
\definecolor{pillD}{HTML}{9C3B7A}   % plum
\definecolor{shadowblue}{HTML}{9EC6F0}

% ---- one uniform type size for every card, well below body size -----------
\newcommand{\csfont}{\comicsans\fontsize{7.0}{7.6}\selectfont}
\newcommand{\csbold}{\comicsans\bfseries\fontsize{7.0}{7.6}\selectfont}
% Comic Sans is reached through WinAnsi, where only the ASCII slots survive, so
% the em dash in the card titles is drawn as a rule (it inherits the text colour).
\newcommand{\csdash}{\,\rule[0.26ex]{0.75em}{0.55pt}\hskip0.3em\relax}

% ---- badge: icon + name on one line, sitting inside the title bar ----------
\newcommand{\casebadge}[3]{% #1 colour, #2 icon, #3 label
  \tikz[baseline=-0.6ex]{\node[inner xsep=2.5pt, inner ysep=1.2pt, fill=white,
      draw=#1!45!black, line width=0.4pt, rounded corners=2pt]
      {\resizebox{!}{7pt}{#2}\hspace{2pt}%
       \raisebox{0.6pt}{\comicsans\bfseries\fontsize{6.0}{6.6}\selectfont
                        \textcolor{#1!80!black}{#3}}};}}

% empty output: a document whose body rows are missing, stamped with a red cross
\newcommand{\iconA}{\tikz[line width=0.5pt]{%
  \draw[fill=white,draw=black!60,rounded corners=1pt] (0,0) rectangle (1.05,1.40);
  \draw[black!40] (0.16,1.14)--(0.89,1.14);
  \draw[black!25,dashed,dash pattern=on 1.2pt off 1.2pt]
        (0.16,0.86)--(0.89,0.86) (0.16,0.58)--(0.89,0.58);
  \fill[caseA] (0.98,0.26) circle (0.38);
  \draw[white,line width=0.8pt] (0.84,0.12)--(1.12,0.40) (1.12,0.12)--(0.84,0.40);}}

% zero dispatch: a lightning bolt struck through
\newcommand{\iconB}{\tikz[line width=0.5pt]{%
  \fill[caseB] (0.62,1.40) -- (0.14,0.62) -- (0.46,0.62) -- (0.30,0.00)
               -- (0.86,0.76) -- (0.52,0.76) -- cycle;
  \draw[caseA!95!black,line width=1.1pt] (-0.02,0.06)--(1.02,1.34);}}

% capacity overflow: a rack filled to the brim with one block pushed out
\newcommand{\iconC}{\tikz[line width=0.5pt]{%
  \draw[fill=white,draw=black!60,rounded corners=1pt] (0.10,0) rectangle (1.00,0.88);
  \fill[caseC!25] (0.19,0.07) rectangle (0.91,0.29);
  \fill[caseC!55] (0.19,0.33) rectangle (0.91,0.55);
  \fill[caseC!85] (0.19,0.59) rectangle (0.91,0.81);
  \fill[caseC] (0.32,1.06) rectangle (0.78,1.30);
  \draw[->,caseC!70!black,line width=0.8pt] (1.20,0.32)--(1.20,1.28);}}

% deadline miss: a clock past its due time
\newcommand{\iconD}{\tikz[line width=0.5pt]{%
  \draw[fill=white,draw=caseD!60!black,line width=0.8pt] (0.66,0.70) circle (0.66);
  \draw[caseD!55!black,line width=0.8pt] (0.66,0.70)--(0.66,1.14);
  \draw[caseD!55!black,line width=0.8pt] (0.66,0.70)--(1.00,0.52);
  \draw[caseA,line width=1.1pt] (1.62,0.42)--(1.62,1.24);
  \fill[caseA] (1.62,0.20) circle (0.11);}}

% ---- the card ------------------------------------------------------------
% #1 case colour, #2 pill colour, #3 title, #4 badge, #5 equal-height group
\newtcolorbox{casecard}[5]{%
  enhanced, width=\linewidth, equal height group=#5,
  colback=white, colframe=#1!72!black, boxrule=0.8pt,
  arc=4pt, outer arc=4pt,
  fuzzy shadow={2pt}{-2pt}{0pt}{0.35mm}{shadowblue!85!white},
  left=3pt, right=3pt, top=1pt, bottom=0.6pt,
  toptitle=1.6pt, bottomtitle=1.6pt,
  colbacktitle=#1, coltitle=white,
  fonttitle=\comicsans\bfseries\fontsize{6.8}{7.5}\selectfont,
  title={\parbox[t]{\dimexpr\linewidth-61pt\relax}{#3}},
  fontupper=\csfont,
  overlay={\node[anchor=east,inner sep=0pt] at ([xshift=-3.5pt]title.east) {#4};}}

% one-line diagnosis pill
\newcommand{\casediag}[3]{% #1 pill colour, #2 number, #3 text
  \noindent\tcbox[on line, colback=#1, colframe=#1!70!black, boxrule=0.4pt,
    arc=2.5pt, left=2.5pt, right=2.5pt, top=0.7pt, bottom=0.7pt, boxsep=0pt,
    coltext=white, fontupper=\csbold]{(#2)\ #3}\par
  \vspace{1.2pt}}

% the highlighted "Symptom." tag, in a snug rounded box
\newcommand{\casesym}[1]{%
  \noindent\tcbox[on line, colback=yellow!62, colframe=yellow!70!orange,
    boxrule=0.4pt, arc=2pt, left=2pt, right=2pt, top=0.7pt, bottom=0.7pt,
    boxsep=0pt, fontupper=\csbold]{Symptom.}%
  \ #1\par}

% verbatim excerpt from the stored answer
\newcommand{\caseexcerpt}[1]{%
  \vspace{1.2pt}\noindent
  \begin{tcolorbox}[enhanced, on line=false, width=\linewidth,
    colback=black!5, colframe=black!25, boxrule=0.4pt, arc=2.5pt,
    left=3.5pt, right=2.5pt, top=1.2pt, bottom=1.2pt, boxsep=0pt,
    before skip=0pt, after skip=1.2pt,
    fontupper=\ttfamily\fontsize{6.5}{7.1}\selectfont]#1\end{tcolorbox}}

% closing analysis, tinted with the case colour
\newcommand{\caseanalysis}[2]{%
  \noindent
  \begin{tcolorbox}[enhanced, on line=false, width=\linewidth,
    colback=#1!7, colframe=#1!55, boxrule=0.5pt, arc=2.5pt,
    left=3.5pt, right=2.5pt, top=1.1pt, bottom=1.1pt, boxsep=0pt,
    before skip=0pt, after skip=0pt,
    fontupper=\csfont]#2\end{tcolorbox}}

\newcommand{\caseprov}[1]{%
  {\csfont\color{black!60}\raggedright #1\par}\vspace{0.8pt}}

% ============================================================================
\begin{figure*}[!t]
\centering

\begin{minipage}[t]{0.487\textwidth}\vspace{0pt}
\begin{casecard}{caseA}{pillA}{Case 1 (Format) \csdash empty dispatch block}{%
  \casebadge{caseA}{\iconA}{FormatVoid}}{fig6top}
\casediag{pillA}{1}{Output-contract breakdown: no generator rows}
\caseprov{\textbf{Falcon3-3B}, untrained ~|~ \texttt{m01d15-rho01-s10}}
\casesym{A \texttt{POWER:} header with nothing under it.}
\caseexcerpt{POWER:\ \ \textrm{\textit{(no generator rows emitted)}}}
\caseanalysis{caseA}{The audit reports omitted generator rows in 37/40 answers and 111 of
  120 missing dispatch values. If all rollouts fail parsing, their identical format penalties
  provide no group-relative learning signal.}
\end{casecard}
\end{minipage}
\hfill
\begin{minipage}[t]{0.487\textwidth}\vspace{0pt}
\begin{casecard}{caseB}{pillB}{Case 2 (Grid) \csdash zero settings for 24\,h}{%
  \casebadge{caseB}{\iconB}{ZeroDispatch}}{fig6top}
\casediag{pillB}{2}{Insufficient generation: demand remains unserved}
\caseprov{\textbf{Falcon3-3B}, untrained ~|~ \texttt{m01d15-rho01-s9}}
\casesym{All dispatch settings are zero throughout the horizon.}
\caseexcerpt{123\_STEAM\_3: 0, 0, 0, \ldots, 0\\ 218\_CC\_1:\ \ \ 0, 0, 0, \ldots, 0}
\caseanalysis{caseB}{Zero settings denote minimum output for online units. This schedule
  leaves 2,825\,MWh unserved; no task meets the grid-aware service criterion.}
\end{casecard}
\end{minipage}

\vspace{2.5pt}

\begin{minipage}[t]{0.487\textwidth}\vspace{0pt}
\begin{casecard}{caseC}{pillC}{Case 3 (Coupling) \csdash data-center overflow}{%
  \casebadge{caseC}{\iconC}{CapOverflow}}{fig6bot}
\casediag{pillC}{3}{Coupling violation: rack capacity \textit{C} exceeded}
\caseprov{\textbf{GPT-5.5}, \textbf{DeepSeek-V4-Flash}, \textbf{Kimi-K2.6}
  ~|~ \texttt{m01d15-rho06-s10}}
\casesym{The output includes a time-varying dispatch,}
\caseexcerpt{123\_STEAM\_3: 50,55,45,50,80,90,90,70, \ldots ,100,95,80,70}
\caseanalysis{caseC}{yet GPT-5.5 puts 12 racks into DC~1 at hour~5 (\textit{C}\,=\,11), and
  DeepSeek-V4-Flash and Kimi-K2.6 each put 14 into DC~2 at hour~20: same constraint,
  same instance, three models.}
\end{casecard}
\end{minipage}
\hfill
\begin{minipage}[t]{0.487\textwidth}\vspace{0pt}
\begin{casecard}{caseD}{pillD}{Case 4 (Service) \csdash deadline \& migratability}{%
  \casebadge{caseD}{\iconD}{DeadlineMiss}}{fig6bot}
\casediag{pillD}{4}{Task-service violation: due time and home DC ignored}
\caseprov{\textbf{Falcon3-3B}, \textbf{Llama-3.2-3B}, untrained ~|~
  \texttt{m01d15-rho01-s9}}
\casesym{Task~11 is released at hour~8 and due at~14 with duration~2.}
\caseexcerpt{11: dc=1 start=14\ \ \ \# ends at 16 > deadline 14}
\caseanalysis{caseD}{Llama-3.2-3B also moves 16 non-migratable tasks away from their home
  data center. These placements violate task constraints and must be excluded from satisfaction.}
\end{casecard}
\end{minipage}

\caption{Failure cases on ECBench, quoted verbatim from the stored model answers.}
\label{fig:cases}
  \Description{Four cards, each quoting a failing model answer with its symptom and diagnosis: an empty dispatch block, an all-zero dispatch, a data-center capacity overflow, and a deadline and migratability violation.}
\end{figure*}

\begin{table*}[!t]
\centering
\fontsize{9}{10}\selectfont
\setlength{\tabcolsep}{4pt}
\caption{Falcon3-3B archived stage comparisons on the same 40 instances, using grid-aware satisfaction (\%) and re-scored cost. The two SFT archives have distinct initialization histories; these are not budget-matched reward ablations.}
\label{tab:stage}
\begin{tabularx}{\textwidth}{@{}>{\hsize=1.6\hsize\linewidth=\hsize\centering\arraybackslash}X>{\hsize=0.88\hsize\linewidth=\hsize\centering\arraybackslash}X>{\hsize=0.88\hsize\linewidth=\hsize\centering\arraybackslash}X>{\hsize=0.88\hsize\linewidth=\hsize\centering\arraybackslash}X>{\hsize=0.88\hsize\linewidth=\hsize\centering\arraybackslash}X>{\hsize=0.88\hsize\linewidth=\hsize\centering\arraybackslash}X@{}}
\toprule
\textbf{Configuration} & \textbf{Viol (MWh)}\,$\downarrow$
 & \textbf{S-cnt}\,$\uparrow$ & \textbf{S-val}\,$\uparrow$ & \textbf{S-rh}\,$\uparrow$
 & \textbf{Cost (\$)}\,$\downarrow$ \\
\midrule
\rowcolor{resultlavender}
PowerAtlas & 216.6 & 69.5 & 65.2 & 62.9 & 3{,}830{,}536 \\
No initialization (archive) & 10{,}730.2 & 3.3 & 3.3 & 3.1 & 98{,}336{,}376 \\
SFT (earlier archive) & 9{,}084.2 & 20.4 & 18.0 & 18.7 & 83{,}257{,}404 \\
SFT-16 (archive) & 682.4 & 43.9 & 43.6 & 39.6 & 7{,}883{,}453 \\
\bottomrule
\end{tabularx}
\end{table*}

\subsection{Residual Constraint Violations (RQ3)}
\label{sec:exp-viol}

\textbf{Dispatch and service failures.} Untrained Falcon3-3B produces usable
dispatches on $3/40$ instances. Figure~\ref{fig:cases} illustrates incomplete
dispatch, supply deficits, capacity overflow, and violations of task deadlines or migration rules.

\textbf{Residual violations.} Training lowers violation six- to fifty-fold
and improves load following (Figure~\ref{fig:fig5}(a)--(b)). Falcon3-3B's
mean/median violation is $216.6/75.5$\,MWh: lost load contributes $73\%$,
over-generation $8\%$, line overload $1\%$, and task violations $17\%$
(rounded). Supply deficits dominate the final LP results. Separately, panel~(f)
shows lower training-side balance residuals on $37/40$ pairs in the earlier diagnostic run.

\textbf{Penalties and feasibility.} Lost-load penalties account for
$65$--$99\%$ of baseline costs and $50$--$66\%$ after training. Remaining
cost gaps reflect physical deficits and task coverage. No selected backbone has
a strictly valid output passing all hourly grid/task checks among its 40 cases
(Appendix~\ref{app:reaudit}); mean improvements do not imply complete feasibility.

\subsection{Checkpoint and Operating-Condition Sensitivity (RQ4)}
\label{sec:exp-sat}

\textbf{Checkpoint sensitivity.} In a separate archived run, initialization raises Falcon3-3B's placement
value satisfaction from $6.8$ to $91.6$ without materially changing the median
training-side balance residual; alignment halves the residual
(Figure~\ref{fig:fig5}(c); Appendix~\ref{app:curves}). Satisfaction peaks at
$93.9$ at step $40$, versus $91.5$ at step $60$ (panel~(d)); selection and reporting share one pool.

\textbf{Operating conditions.} Trained value satisfaction spans
$89.8$--$97.3$ across sampled loads and seasons, versus below $25\%$ before
training (Figure~\ref{fig:fig5}(e)). These results characterize variation within
shared training/reporting dates and load regimes. Transfer to unseen conditions,
topologies, or horizons remains to be evaluated using held-out regimes and
independent checkpoint selection.

\textbf{Service trade-offs.} In the paired Qwen subset, grid-aware value
satisfaction is $79.0\%$, but low-priority service is $3.1\%$ versus $91.8\%$
for high-priority tasks. The corresponding low-priority rates are $68.3\%$
for Llama and $66.9\%$ for Falcon (Table~\ref{tab:backbone}). These tier-specific
outcomes complement aggregate value satisfaction.

\section{Conclusion}
\label{sec:conclusion}
PowerAtlas studies verifier-guided structured generation for electricity--computing
co-scheduling using ECBench, with 2,000 instances and solved references.
On identical archived instances for each backbone, selected checkpoints reduce
mean violation by 89--98\% and penalized cost by up to 26$\times$ relative to base
models. Stage comparisons use unified grid-aware scoring. Residual physical violations and strict-output
failures remain: these results demonstrate improved retrospective scores, while
independent testing and complete schedule verification remain necessary for
establishing reliable co-scheduling performance.

\FloatBarrier
\clearpage
\section*{Limitations}
\label{sec:limitations}

Reported gains across three backbones do not establish consistent feasibility:
Falcon3-3B retains $216.6$\,MWh mean violation, largely unserved load.
Shared operating regimes, checkpoint selection, and incomplete configuration provenance
\mbox{also limit generalization claims.}

Future research will evaluate all methods on the full test set with consistent metrics
and independent checkpoint selection. It will test held-out operating regimes and compare
solver and retrieval baselines under matched \mbox{settings and latency budgets.}

\section*{Ethics Statement}
\label{sec:ethics}

The reported experiments evaluate electricity--computing co-scheduling in a
modeled benchmark. Generated schedules can violate grid limits
or leave computing tasks without power, even when aggregate scores improve.
Operational use would require independent feasibility checks, human oversight,
and safe fallback procedures; the reported results do not establish deployment
readiness. Value-weighted scheduling can also disadvantage low-priority tasks.
Evaluation should therefore report service outcomes across priority tiers,
rather than rely only on aggregate value or cost.

ECBench combines a power-system test model with synthetic computing tasks
calibrated from facility records. Sharing and reuse of research artifacts
should protect sensitive facility information and respect the licenses and
access conditions of source data and model backbones. The public repository
supports research transparency; benchmark results should not be interpreted
as guarantees of safety, fairness, or reliability in operational settings.

\section*{Acknowledgments}
This work was supported by the Fundamental Research Funds for the Beijing
University of Posts and Telecommunications (Grant 2025TSQY03). We thank a
provincial electric power company in China for jointly building the experimental
network and providing scoped data-center access without disclosing
security-sensitive operational data.

\bibliography{references/references}

\clearpage
\appendix

\newsavebox{\appendixpromptbox}
\begin{lrbox}{\appendixpromptbox}
\begin{minipage}{\textwidth}
\section{Prompts Used in PowerAtlas}
\label{app:prompts}
PowerAtlas serializes each instance into a structured prompt and generates the joint decision
as one templated response. Figure~\ref{fig:app-prompts} presents the scheduling interface~(a),
retrieval query~(b), and an illustrative supervised target~(c). The verifier and reward computation are given in Appendix~\ref{app:reward}.
These templates describe the interface, rather than verbatim transcripts of every
historical run. Placeholders in \texttt{\{curly braces\}} are filled per instance using the corresponding grid and task specification.

\begingroup
\centering
\begin{promptcard}{(a)\quad ECCS Scheduling Prompt}
\begin{lstlisting}[style=promptin]
[SYSTEM]
You are an electricity-computing co-scheduling agent. Given a 24-hour power-system state and a
list of computing tasks, output a single schedule that (1) respects power balance, DC power-flow,
and line-capacity limits; (2) honors per-data-center power capacity and each task's release,
deadline, and tier; and (3) minimizes total operating cost. First reason inside <reason>...
</reason>, then output ONLY the decision inside <answer>...</answer>.

[USER]
Grid: RTS-GMLC, {n_bus} buses, {n_line} lines, {n_unit} thermal units.
Controllable generators (the swing unit absorbs the residual): {unit_table}
Hourly signal h=1..24: load / renewable / net_load / fixed_gen = {profiles}
Data centers: {n_dc} DCs, each with capacity C={C} racks, peak power {Pmax_dc} MW,
              idle power {Pidle} MW.
Tasks (one row each):
  id | origin_dc | release | deadline | duration | demand | power | migratable |
     priority | drop_penalty | mig_cost
{task_rows}
Retrieved solved cases (optional priors): {retrieved_cases}

CONSTRAINTS
  - a placed task starts within [release, deadline - duration + 1];
  - a non-migratable task runs at its origin DC; a migratable one may run at any DC;
  - at every DC and hour: active rack-demand <= C, and DC load (idle + task power)
    <= DC peak power;
  - generation (fixed + controllable + swing) must cover the grid background plus all
    data-center power, and each controllable unit must stay within [Pmin, Pmax].
  (task 'demand' = number of racks; task 'power' = MW draw.)

[OUTPUT FORMAT]
<reason> ... </reason>
<answer>
TASKS: <per task>  drop  |  (dc=<id>, start=<hour>)
POWER: <per controllable unit>  24 hourly levels, each 'off' or a number in [0,100]
       meaning percent between Pmin and Pmax
</answer>
\end{lstlisting}
\end{promptcard}

\begin{promptcard}{(b)\quad Knowledge Retrieval Query}
\begin{lstlisting}[style=promptin]
[RETRIEVAL QUERY]
Season={season}; peak_load={peak}; renewable_share={rshare}; congestion={congested_lines};
task_mix={tier_histogram}; horizon=24h.
Return the top-{k} training instances whose (grid regime, task profile) are most similar, each
with its Gurobi-optimal TASKS/POWER decision as an in-context prior.
\end{lstlisting}
\end{promptcard}

\begin{promptcard}{(c)\quad Supervised Initialization Target}
\begin{lstlisting}[style=promptin]
<reason>
Peak net load at h=18; lines {L12,L27} near limit => pre-commit units {g2,g5};
defer low-tier tasks {t7,t9} to off-peak; place high-tier tasks {t1,t3} at release.
</reason>
<answer>
TASKS: t1 (dc=2, start=9); t2 drop; t3 (dc=1, start=14); ...
POWER: g1 [p1..p24]; g2 [p1..p24]; ...; g5 [p1..p24]   # percent of Pmin..Pmax
</answer>
\end{lstlisting}
\end{promptcard}
\captionof{figure}{PowerAtlas interface templates: instance serialization and output contract~(a),
a retrieval query~(b), and an illustrative supervised target~(c). Retrieval is used during training only.}
\label{fig:app-prompts}
  \Description{Three boxed interface templates: the ECCS scheduling prompt, the knowledge retrieval query, and the supervised initialization target.}
\endgroup

\end{minipage}
\end{lrbox}
\twocolumn[\usebox{\appendixpromptbox}]

\subsection{ECCS Scheduling Prompt} The agent receives the grid signal, the task
list, and (optionally) retrieved solved cases, and must emit a two-block decision: one line per
task and one $24$-point trajectory per agent-controlled unit.
The verifier scores only the content of \texttt{<answer>}.

\paragraph{Supervised target.} During supervised initialization, the policy is trained to reproduce the templated serialization of the oracle decision $D^{\star}$. The
\texttt{<reason>} field is a short rationale distilled from the solved instance, while the verifier evaluates only the \texttt{<answer>} block.

\subsection{Knowledge Retrieval Prompt} Solved training cases are retrieved with
the hypergraph retriever of Graph-R1~\cite{Luo2026GraphR1,Luo2025HyperGraphRAG}; the query encodes
the instance's grid regime and task profile so that priors match the current state.

\FloatBarrier

\section{Theoretical Proof}
\label{app:proof}
We establish three properties of the reward, group-relative update, and hourly evaluator. Each follows from the stated equations under explicit assumptions. The statements concern the scoring equations and evaluator under these assumptions.

\subsection{Proof of Proposition 1}
\label{app:proof-reward}
\paragraph{Proposition 1.}
Suppose $\kappa,\phi,s_{\mathrm{val}},s_{\mathrm{cnt}}\in[0,1]$, the weights are nonnegative with $\eta_v+\eta_c=w_c+w_p=1$, and $c^{\star}_{\mathrm{tr}}\geq0$, $\lambda_{\mathrm{tr}}\geq0$. Wherever $c^{\mathrm{tr}}_{\theta}+\lambda_{\mathrm{tr}}V_{\mathrm{tr}}>0$, the well-formed base reward in Equations~\eqref{eq:rquality}--\eqref{eq:rgate} lies in $[0,1]$. With cost, reference cost, and task-quality terms fixed, this reward is nonincreasing in $V_{\mathrm{tr}}$.

\paragraph{Proof.}
The retained-value/count mixture is a convex combination of two numbers in $[0,1]$. Multiplication by $\kappa\phi\in[0,1]$ preserves that interval, so $Q_{\mathrm{compute}}\in[0,1]$. The grid ratio is nonnegative, and clipping it above at one gives $Q_{\mathrm{grid}}\in[0,1]$. Their convex combination therefore lies in $[0,1]$.
For $V_2\geq V_1$ with positive denominators,
\begin{equation}
\frac{c^{\star}_{\mathrm{tr}}}{c^{\mathrm{tr}}_{\theta}+\lambda_{\mathrm{tr}}V_2}
\leq
\frac{c^{\star}_{\mathrm{tr}}}{c^{\mathrm{tr}}_{\theta}+\lambda_{\mathrm{tr}}V_1}.
\end{equation}
Taking the minimum with one preserves this ordering, as does multiplication by $w_p\geq0$ and addition of the fixed computing term. On the unsaturated branch,
\begin{equation}
\frac{\partial Q_{\mathrm{grid}}}{\partial V_{\mathrm{tr}}}
=-\frac{\lambda_{\mathrm{tr}}c^{\star}_{\mathrm{tr}}}
{(c^{\mathrm{tr}}_{\theta}+\lambda_{\mathrm{tr}}V_{\mathrm{tr}})^2}\leq0.
\end{equation}
The malformed-output reward is $-r_0<0$, below every well-formed base reward.\hfill$\square$

\paragraph{Consequence.}
The power term penalizes larger training residuals when the other terms are held fixed. Across different schedules, cost and task quality can also change, so a larger total reward need not imply a smaller violation. The bound applies to the base reward; the available implementation's capped retrieval bonus is documented in Appendix~\ref{app:impl}.

\subsection{Proof of Proposition 2}
\label{app:proof-advantage}
\paragraph{Proposition 2.}
For a group of $N\geq2$ finite rewards, let $\bar R$ be their mean, $s_R$ their standard deviation, and $\delta>0$. The advantages in Equation~\eqref{eq:adv} sum to zero. If all rewards are identical, every advantage is zero and the reward-driven clipped surrogate in Equation~\eqref{eq:grpo} contributes no policy gradient for that group.

\paragraph{Proof.}
All advantages have the same positive denominator. Thus
\begin{equation}
\sum_{i=1}^{N}\hat A_i
=\frac{\sum_i R_i-N\bar R}{s_R+\delta}=0.
\end{equation}
If $R_i=r$ for every $i$, then $\bar R=r$, $s_R=0$, and every numerator vanishes. Treating sampled rewards and advantages as fixed during the policy update, both arguments of the clipped minimum are then zero for every token and every policy ratio. Their contribution and its gradient are zero. The KL regularizer is separate and may still contribute a gradient.\hfill$\square$

\paragraph{Consequence.}
A group consisting entirely of malformed outputs receives the same $-r_0$ penalty and supplies no relative reward signal. This explains the role of supervised initialization in obtaining parseable trajectories before alignment.

\subsection{Proof of Proposition 3}
\label{app:proof-lp}
\paragraph{Proposition 3.}
Assume the hourly LP in Equation~\eqref{eq:eval-lp} is feasible and attains its finite optimum. Its optimal objective is zero if and only if its fixed dispatch and task-induced loads admit a solution with zero lost-load, over-generation, and line-overload slacks, using the evaluator's allowed swing dispatch and renewable curtailment. Zero objectives at every hour do not certify all intertemporal ECCS constraints.

\paragraph{Proof.}
The objective is the sum of nonnegative slacks with strictly positive unit coefficients. A zero optimum therefore requires every slack in an optimal solution to be zero. Conversely, any feasible solution with all those slacks zero attains objective zero, which is globally minimal by nonnegativity.
For the second statement, consider an illustrative two-hour, single-bus system with a fixed non-swing generator, output bounds $[0,100]$ MW, an upward ramp limit of $10$ MW/h, and no renewable or swing contribution. Set dispatch and demand to $20$ MW in hour one and $80$ MW in hour two. Each hourly balance is exact and there are no lines to overload, so both hourly objectives are zero. However, the $60$ MW increase exceeds the ramp limit. Thus hourly feasibility does not imply feasibility of the joint two-hour schedule.\hfill$\square$

\paragraph{Consequence.}
The evaluator certifies only the constraints it includes, with the swing unit re-optimized at each hour. Ramping, commitment, and task legality require their corresponding checks; coverage is listed in Appendix~\ref{app:coverage}. A zero hourly grid objective should not be interpreted as a guarantee of complete ECCS feasibility.

\FloatBarrier

\section{PowerAtlas Algorithm Details}
\label{app:impl}
Algorithm~\ref{alg:poweratlas-workflow} connects dataset preparation, the two training stages, and evaluation. Algorithm~\ref{alg:fagrpo} gives the group-relative update itself. Retrieval is available in training rollouts, while reported inference generates one joint decision without retrieved examples, solver calls, or repair.

\begin{algorithm}[!t]
\footnotesize
\caption{PowerAtlas Training and Evaluation Workflow}
\label{alg:poweratlas-workflow}
\begin{algorithmic}[1]
\Require Solved training instances $\mathcal D_{\mathrm{train}}$; pretrained policy; reporting pool $\mathcal E$; training configuration.
\Ensure Trained policy and metrics for its generated decisions.
\State Serialize training instances and oracle targets using Appendix~\ref{app:prompts}.
\State Build the retrieval index from training cases and expert notes (Appendix~\ref{app:corpus}).
\State Run supervised initialization with Equation~\eqref{eq:sft}; freeze the resulting reference policy.
\For{each configured alignment step}
  \State Sample an instance batch and $N$ trajectories per instance, with training-time retrieval where enabled.
  \State Parse answers and compute training rewards; record the scorer version, penalty, and any retrieval bonus.
  \State Compute group-relative advantages using Equation~\eqref{eq:adv}.
  \State Update with the clipped objective and KL regularization in Equation~\eqref{eq:grpo}.
\EndFor
\State Select the checkpoint under the reported protocol (Appendix~\ref{app:protocol}).
\For{each instance in $\mathcal E$}
  \State Generate one greedy response without retrieval, solver calls, or repair.
  \State Parse the joint task and power decision.
  \State Apply task checks and the hourly nodal LP to the submitted decision; record metrics and latency.
\EndFor
\end{algorithmic}
\end{algorithm}

\paragraph{Output parsing and reward.}
\label{app:reward} The training scorer parses the
\texttt{<answer>} block and computes task quality and a power score. Its available
implementation uses aggregate balance rather than the hourly nodal DC LP employed
for final evaluation. Section~\ref{sec:method-grpo} defines the base reward; malformed
outputs receive $-r_0$. The count weight $\eta_c$ reduces the incentive to drop low-value
tasks, while the power penalty discourages leaving demand uncovered. Run-specific
settings and retrieval bonuses are discussed in Appendix~\ref{app:impl}.

\begin{figure}[!t]
\begin{promptcard}{Verifier and Trajectory Reward}
\begin{lstlisting}[style=promptin]
parse(<answer>) -> {TASKS[], POWER[24 x n_controlled]}
  if not well_formed:            return reward = -r0            # r0 = 0.5
  # --- computing side -------------------------------------------------
  retained  = eta_v * s_val + eta_c * s_cnt  # fractions in [0,1]; 0.7 / 0.3
  Q_compute = retained * coverage * (1 if task_checks_pass else 0.5)
  # --- training-side power score (not the final nodal LP) -------------
  V_train = aggregate_balance_residual(POWER, TASKS)
  Q_grid  = min(1, c_train_ref/(c_train_policy + lambda_train * V_train))
  # --- base reward ---------------------------------------------------
  reward = w_c * Q_compute + w_p * Q_grid
  # Available code additionally applies a capped retrieval bonus.
  # Historical run-level settings require configuration reconciliation.
\end{lstlisting}
\end{promptcard}
\end{figure}

\paragraph{Complexity Analysis.}
Let $B$ be the number of instances in a training batch, $N$ the trajectories per instance, $T_{\mathrm{roll}}$ the cost of generating one trajectory, and $T_{\mathrm{score}}$ its scoring cost. The batch requires $BN$ rollouts and scorer calls; their serial work is proportional to $BN(T_{\mathrm{roll}}+T_{\mathrm{score}})$, in addition to retrieval and the policy update. Parallel serving changes wall-clock time without changing these counts. Table~\ref{tab:hparams} specifies the group size and generation budgets.
At inference, the policy generates one response. The separate evaluator solves $H=24$ hourly feasibility programs and applies task checks after generation. The one-response interface avoids repeated policy--solver interaction. Wall-clock latency depends on hardware, response length, and serving conditions.

\paragraph{Hyperparameters.}
Table~\ref{tab:hparams} records the reported settings for the three backbones,
with model-specific exceptions shown explicitly.
Full-parameter supervised initialization is followed by critic-free policy alignment with vLLM rollouts.
Inference uses greedy autoregressive decoding in one response, without a solver.

Training rollouts may interleave retrieval and generation for up to $8$ turns, returning
the top-$5$ entries per retrieval with a total response budget of $4{,}096$ tokens.
Tool calls within a turn run in parallel. With $N=2$ trajectories and batches of $8$ instances, training fits on two GPUs.
Each group-relative advantage is therefore based on only one pair of rewards.
Identical rewards yield zero relative advantage. The grid term provides graded feedback
for parseable schedules; malformed outputs receive the fixed penalty $-r_0$.
Supervised initialization helps avoid groups in which every output receives this same penalty.

\begin{table}[!t]
\centering
\caption{Reported training and inference settings. The available scorer differs
from the nominal training penalty; see the implementation-provenance note below.}
\label{tab:hparams}
\tablefont
\setlength{\tabcolsep}{3pt}
\begin{tabularx}{\columnwidth}{@{}>{\raggedright\arraybackslash}p{0.40\columnwidth}>{\raggedright\arraybackslash}X@{}}
\toprule
\textbf{Setting} & \textbf{Value} \\
\midrule
Backbones & Qwen3-4B, Llama-3.2-3B, Falcon3-3B \\
Initialization & full-parameter SFT \\
SFT epochs / lr / batch & $8$ ($16$ for Falcon3-3B) / $1\!\times\!10^{-5}$ / $16$ \\
SFT max sequence length & $4{,}096$ tokens \\
alignment steps & $60$ (shipped: step $40$; step $20$ for Llama-3.2-3B) \\
Advantage estimator & Group-relative (critic-free) \\
Group size $N$ & $2$ \\
KL weight $\beta$ / clip $\epsilon$ & $1\!\times\!10^{-3}$ (low-variance KL) / $0.2$ \\
Learning rate / batch size & $2\!\times\!10^{-6}$ / $8$ \\
Entropy coefficient & $0$ \\
Prompt / response length & $8{,}192$ / $2{,}048$ tokens per turn \\
Retrieval top-$k$ & $5$ \\
Reward weights $(w_c,w_p)$ & $(0.5,\,0.5)$ \\
Value/count weights $(\eta_v,\eta_c)$ & $(0.7,\,0.3)$ \\
Infeasibility discount & $0.5$ \\
Evaluation $C_{\mathrm{LOL}}$ / format penalty $r_0$ & $9{,}000$\,\$/MWh / $0.5$ \\
Rollout engine / inference decoding & vLLM $0.8.4$ / greedy \\
Hardware & $2\times$ NVIDIA A800 ($80$\,GB) \\
\bottomrule
\end{tabularx}
\end{table}

\paragraph{Implementation provenance.}
The available fast scorer sets $\lambda_{\mathrm{tr}}=3{,}000$\,\$/MWh, whereas the
nominal configuration previously reported $9{,}000$. It also adds
$0.1\min(n_{\mathrm{search}},2)$ to the base reward for a well-formed output.
The retained code snapshot does not identify the scorer configuration and rollout
sampling settings for each historical checkpoint. Table~\ref{tab:hparams} records
the reported settings; greedy decoding refers to inference.

\paragraph{Reproducibility.}
The local implementation uses PyTorch $2.6.0$ (CUDA $12.4$), vLLM $0.8.4$,
Transformers $4.51.3$, FlashAttention $2.7.4$, and Gurobi $13.0.2$.
Initialization uses one GPU; Policy alignment uses two A800s with FSDP and reference-policy parameter
offloading. vLLM uses a GPU memory utilization setting of $0.6$.
The retrieval index is local; frontier-model evaluation uses remote APIs.
Seed-based split membership can be recovered from instance identifiers
(\texttt{m<month>d<day>-rho<level>-s<seed>}).
The code and data repository is linked in the abstract.
Reported step times are tens of seconds for initialization and about a minute for policy alignment,
including rollouts; total runtime depends on the number of steps and GPUs used for the corresponding training run.

\FloatBarrier

\section{ECBench Detail}
\label{app:ecbench}
This appendix expands Section~\ref{sec:bench-design} into a self-contained reference: the task and decision interface (Appendix~\ref{app:spec}), benchmark construction (Appendix~\ref{app:construction}), and the evaluator and metric definitions (Appendix~\ref{app:evallp}).

\subsection{Task Design}
\label{app:spec}
The input specification $S=(S_C,S_T,S_K,S_F)$ comprises grid physical constraints $S_C$, the task
description $S_T$, domain knowledge $S_K$, and the output-format contract $S_F$
(Figure~\ref{fig:task}). Retrieval augments training only; reported inference contains
no retrieved cases (Appendix~\ref{app:corpus}). Table~\ref{tab:ecbench-stats} summarizes the key parameters.

\begin{table}[!t]
\centering
\caption{Key parameters of an ECBench instance.}
\label{tab:ecbench-stats}
\tablefont
\begin{tabularx}{\columnwidth}{@{}>{\raggedright\arraybackslash}p{0.43\columnwidth}>{\raggedright\arraybackslash}X@{}}
\toprule
\textbf{Component} & \textbf{Value} \\
\midrule
Buses / lines / thermal units & $73$ / $120$ / $73$ \\
Agent-controlled units & $6$ (of which $1$ swing) \\
Economic-dispatch units & $67$ \\
Scheduling horizon & $24$ hours \\
Scenario grid & $50$ days $\times$ $4$ $\rho$ levels $\times$ $10$ seeds \\
Data-center load fraction $\rho$ & $\{1,2,4,6\}\%$ \\
Instances (train / test) & $1{,}600$ / $400$ \\
Oracle solver & Gurobi $13.0.2$ (MILP, exact) \\
Tasks per instance & $37$ (fixed; $74{,}000$ in total) \\
Tier distribution (H/M/L) & $50.8\%$ / $21.9\%$ / $27.3\%$ \\
Task power / duration & $1.4$--$33.5$\,MW / $2$--$6$\,h \\
Retrieval corpus & $551$ entries (train split only) \\
\bottomrule
\end{tabularx}
\end{table}

\paragraph{Optimization model.}
\label{app:milp}
The oracle decision $D^{\star}=(D_P^{\star},D_C^{\star})$ solves the following mixed-integer program. Notation follows Section~\ref{sec:task}, and cost coefficients are specified in the released solver configuration. Let $\mathcal{G}$ be the thermal units, $\mathcal{L}$ the
lines, $\mathcal{N}$ the buses, $\mathcal{T}=\{1,\dots,24\}$ the hours, $\mathcal{K}$ the tasks,
and $\mathcal{D}$ the data centers. The decision variables are the dispatch $p_{g,t}$ and
commitment $u_{g,t}\in\{0,1\}$ of the $6$ agent-controlled units, the task placement
$x_{k,d,t}\in\{0,1\}$ (task $k$ started at hour $t$ in data center $d$), and the drop indicator
$z_k\in\{0,1\}$; the remaining $67$ units follow economic dispatch.
\begin{align}
\min\ \ & \textstyle\sum_{t}\sum_{g} C_g(p_{g,t})\notag\\
&+ \sum_{k}\big(\pi_k m_k + v_k z_k\big)\notag\\
&+ C_{\mathrm{LOL}}\sum_{t}\sum_{n}p^{\downarrow}_{n,t}, \label{eq:obj}\\
\text{s.t.}\ \ & \textstyle\sum_{g} p_{g,t} + \sum_{n}\big(R_{n,t}-c_{n,t}+p^{\downarrow}_{n,t}\big)\notag\\
&\quad= \sum_{n}\big(d_{n,t}+q_{n,t}\big), \ \ \forall t, \label{eq:bal}\\
& 0\le c_{n,t}\le R_{n,t}, \ \ \forall n,t, \label{eq:curt}\\
& \textstyle\sum_{g}\big(\overline P_{g}u_{g,t}-p_{g,t}\big)\ge \gamma\sum_{n} d_{n,t},
        \ \ \forall t, \label{eq:res}\\
& f_{\ell,t}=\big(\theta_{a(\ell),t}-\theta_{b(\ell),t}\big)/x_{\ell},\notag\\
& |f_{\ell,t}|\le \bar F_{\ell}, \ \ \forall \ell,t, \label{eq:dcflow}\\
& \nu_{n,t}+p^{\downarrow}_{n,t}\notag\\
&\quad= \textstyle\sum_{\ell\in\delta^{+}(n)} f_{\ell,t}-\sum_{\ell\in\delta^{-}(n)} f_{\ell,t},
  \ \ \forall n,t, \label{eq:kcl}\\
& u_{g,t}\underline P_{g}\le p_{g,t}\le u_{g,t}\overline P_{g}, \ \ \forall g,t, \label{eq:gen}\\
& |p_{g,t}-p_{g,t-1}|\le R_{g}, \ \ \forall g,t, \label{eq:ramp}\\
& q_{d,t}=\textstyle\sum_{k}\sum_{\tau} e_{k}\,x_{k,d,\tau}\,\notag\\
&\quad \mathbf{1}[\tau\le t<\tau+\Delta_k], \ \ \forall d,t, \label{eq:dcload}\\
& q_{d,t}\le \overline Q_{d}, \notag\\
&\textstyle\sum_{k}\sum_{\tau} r_{k}\,x_{k,d,\tau}\,\notag\\
&\quad \mathbf{1}[\tau\le t<\tau+\Delta_k]\le C, \ \ \forall d,t, \label{eq:cap}\\
& \textstyle\sum_{d}\sum_{t=r_k}^{\bar r_k-\Delta_k+1} x_{k,d,t} + z_k = 1, \ \ \forall k, \label{eq:task}\\
& x_{k,d,t}=0, \ \ \forall t,\ \forall d\neq o_k \ \text{ with } \mu_k=0. \label{eq:mig}
\end{align}
where $\nu_{n,t}=\sum_{g\in n}p_{g,t}+R_{n,t}-c_{n,t}-d_{n,t}-q_{n,t}$ is the nodal injection
(the sum over $g\in n$ includes all units at bus $n$: agent-controlled, swing, and
economic-dispatch units), $\theta_{n,t}$ the bus phase angle with $\theta_{\mathrm{ref},t}=0$,
$x_{\ell}$ the line reactance,
$x_{k,d,t},z_k,u_{g,t}\in\{0,1\}$, $C_g$ is the generation cost, $\pi_k m_k$ a migration penalty,
$v_k$ the drop penalty, $C_{\mathrm{LOL}}$ the value of lost load, $p^{\downarrow}_{t}$ the lost load,
$q_{d,t}$ the data-center power (drawn at its host bus), and $e_k$/$\Delta_k$/$[r_k,\bar r_k]$ the
task power/duration/release--deadline window. Equation~\eqref{eq:bal} enforces system power
balance, \eqref{eq:dcflow} the DC power-flow and line limits, \eqref{eq:gen} and~\eqref{eq:ramp} the
generator operating envelope and ramping, \eqref{eq:dcload} and~\eqref{eq:cap} data-center load coupling and capacity, and \eqref{eq:task} that each task is scheduled once within its window or
dropped, and \eqref{eq:mig} that a non-migratable task ($\mu_k=0$) stays at its home data center
$o_k$. Here $R_{n,t}$ is the available renewable at bus $n$ and $c_{n,t}$ its curtailment,
$p^{\downarrow}_{n,t}\ge 0$ the lost load at bus $n$ (priced at $C_{\mathrm{LOL}}$ in the
objective), $\gamma=0.03$ the spinning-reserve margin, $r_k$ the rack demand of task $k$ and $C$ the
rack capacity of a data center. Minimum up- and down-time constraints on $u_{g,t}$ are enforced in the solved program
in the usual big-$M$ form and are omitted here; the oracle formulation includes them. The hourly evaluator's constraint coverage is specified in Appendix~\ref{app:coverage}.
The oracle permits penalized lost load, so solver optimality does not itself imply
zero unmet demand.

Two evaluator conventions affect the reported metrics. First, the MWh-equivalent
$\mathrm{Viol}_{\mathrm{task}}$ in Equation~\eqref{eq:viol} converts task-side hard
violations into power. For each (data center, hour) cell with rack demand above capacity
$C$, excess racks are converted using the per-rack power draw. For cells with total draw
above $\overline Q_{d}$, excess megawatts are counted directly. Window, migratability,
and invalid-placement violations add no MW term but exclude the task from satisfaction scores. Second, the migration term of~\eqref{eq:obj} is
$\pi_k m_k$ with $m_k\in\{0,1\}$ indicating that task $k$ runs outside $o_k$ and
$\pi_k=8\,r_k$ its price in the released cost configuration. Migration charges are proportional
to the number of racks moved. The program is solved to optimality with Gurobi~\cite{Gurobi2024} to
obtain $D^{\star}$ and the reference cost $c^{\star}$.

\subsection{Benchmark Construction}
\label{app:construction}
Each ECBench instance couples a power-system model with a data-center workload and is paired with
an oracle-optimal decision. \emph{(i) Grid side.} We build on the RTS-GMLC test
system~\cite{Barrows2020RTS} ($73$ buses, $120$ transmission lines, and $73$ thermal units, of
which $6$ are agent-controlled and $67$ follow economic dispatch), and evaluate all combinations of
$50$ representative days spanning all twelve months, four data-center load levels
$\rho\in\{1,2,4,6\}\%$, and $10$ random seeds, to obtain hourly profiles of load, renewable generation, net load, and
aggregate fixed generation over a $24$-hour horizon. One of the six controlled units is
the swing unit, which absorbs residual imbalance. \emph{(ii) Computing side.} Each
instance contains $37$ synthetic computing tasks calibrated from data-center records, each with a
priority tier (high/medium/low), a release time, a deadline, a value, and a drop penalty.
Task power ranges over $1.4$--$33.5$\,MW and duration over $2$--$6$ hours, and the tier mix across the
$74{,}000$ tasks of the benchmark is $50.8\%$ high, $21.9\%$ medium, $27.3\%$ low.
\emph{(iii) Coupling.} Nodal power balance, DC power-flow and line-capacity limits, and per-data-center power
capacity link the two domains: task placement changes nodal demand, and the grid state
constrains admissible placements. \emph{(iv) Oracle labels.} Every instance
is solved to optimality by the Gurobi MILP solver~\cite{Gurobi2024}, yielding the reference
decision $D^{\star}$ and reference cost $c^{\star}$ used for both supervised initialization and
reward normalization. We generate $2{,}000$ instances and split them by random seed into $1{,}600$
training and $400$ test instances. The splits share operating regimes, while the retrieval
corpus uses \emph{only} training instances. Checkpoint selection and reporting-pool membership are specified in Appendix~\ref{app:protocol}.

\paragraph{Retrieval corpus.}
\label{app:corpus}
The corpus contains $551$ entries: $520$ solved ECCS instances with grid regimes, task profiles,
and Gurobi-optimal schedules, plus $31$ expert notes on scheduling and data-center energy use.
All cases come from the $1{,}600$-instance training split.
Entries are indexed using the Graph-R1 hypergraph retriever~\cite{Luo2026GraphR1,Luo2025HyperGraphRAG}.
Retrieval is used during training; reported inference runs contain no retrieved cases in $S_K$ when generating a joint schedule.

\begin{table*}[t]
\centering
\tablefont
\setlength{\tabcolsep}{4pt}
\caption{Constraint coverage of the oracle formulation and available scoring implementations.
Enforced denotes an optimization constraint; checked denotes detection or scoring,
not repair. The training column describes the available fast scorer; matching it to a
particular historical checkpoint requires the corresponding run configuration.}
\label{tab:constraint-coverage}
\begin{tabularx}{\textwidth}{@{}>{\raggedright\arraybackslash}p{0.25\textwidth}
>{\raggedright\arraybackslash}X
>{\raggedright\arraybackslash}X
>{\raggedright\arraybackslash}X@{}}
\toprule
\textbf{Constraint or check} & \textbf{Oracle MILP} & \textbf{Training-side fast scorer} & \textbf{Final hourly evaluator}\\
\midrule
Task release, deadline, and migration eligibility
& Enforced by admissible placements & Checked & Checked\\
Rack capacity and data-center power
& Enforced & Checked & Checked; excess contributes to task violation\\
Power balance
& Nodal balance with lost-load slack & Aggregate balance residual & Nodal LP with lost-load and over-generation slacks\\
DC power-flow equations
& Enforced & Not checked & Enforced in hourly LP\\
Transmission capacity
& Enforced & Not checked & Overload slack is measured\\
Non-swing generator output range
& On/off and operating bounds & Percentage-to-MW mapping & Percentage-to-MW mapping\\
Swing-unit operating range
& On/off and operating bounds & Off or within operating bounds & Continuous range from zero to maximum output\\
Generator ramping
& Enforced & Not checked & Not checked\\
Minimum up/down times
& Enforced & Not checked & Not checked\\
Spinning reserve
& Enforced & Not checked & Not checked\\
Missing dispatch units or hours
& Not applicable & Flagged; missing levels mapped to zero output & Mapped to zero output\\
\bottomrule
\end{tabularx}
\end{table*}

\subsection{Evaluation Metric Design}
\paragraph{Hourly feasibility program.}
\label{app:evallp}
The evaluator computes $\mathrm{Viol}$ independently of the model's stated calculations.
For each hour $t$, it fixes the dispatch of the five non-swing controlled units and the
data-center load implied by task placements, then solves the nodal DC feasibility LP
\begin{equation}
\label{eq:eval-lp}
\begin{aligned}
\min\ \ & \textstyle\sum_{\ell\in\mathcal{L}} f^{+}_{\ell,t}
          + \sum_{n\in\mathcal{N}}\big(p^{\downarrow}_{n,t}+p^{\uparrow}_{n,t}\big) \\
\text{s.t.}\ \ & \nu_{n,t}+p^{\downarrow}_{n,t}-p^{\uparrow}_{n,t}\\
&\quad= \textstyle\sum_{\ell\in\delta^{+}(n)} f_{\ell,t}-\sum_{\ell\in\delta^{-}(n)} f_{\ell,t}, \\
& \nu_{n,t}=g^{\mathrm{fix}}_{n,t}+g^{\mathrm{ctrl}}_{n,t}+g^{\mathrm{sw}}_{n,t}\\
&\qquad+R_{n,t}-c_{n,t}-d_{n,t}-q_{n,t}, \\
& f_{\ell,t}=\big(\theta_{a(\ell),t}-\theta_{b(\ell),t}\big)/x_{\ell},\\
& |f_{\ell,t}|\le \bar F_{\ell}+f^{+}_{\ell,t}, \\
& \theta_{\mathrm{ref},t}=0,\quad 0\le c_{n,t}\le R_{n,t},\\
& 0\le g^{\mathrm{sw}}_{t}\le \overline P_{\mathrm{sw}},
\end{aligned}
\end{equation}
with all slacks non-negative, where $\nu_{n,t}$ is the nodal injection, $R_{n,t}$ the available
renewable at bus $n$ and $c_{n,t}$ its curtailment, and $p^{\downarrow}$, $p^{\uparrow}$ and $f^{+}$
are the lost-load, over-generation and line-overload slacks of Section~\ref{sec:bench-design}. The evaluator re-optimizes the swing unit rather than using a model-specified dispatch. Its capacity limits determine how much imbalance it can absorb.
Summing the three types of slack over $24$ hours gives the grid contribution to
$\mathrm{Viol}$; task violations are added as in Equation~\eqref{eq:viol}.
The same LP supplies Figure~\ref{fig:fig5}(b).
Non-negative slacks permit scoring schedules with unmet demand or line overloads.
These hourly checks do not certify full intertemporal feasibility.

The evaluator of Section~\ref{sec:bench-design} reports violations and penalized cost.
\textbf{Viol} combines grid slacks with task-capacity penalties in MWh-equivalent units;
\textbf{S-cnt}/\textbf{S-val}/\textbf{S-rh} are the credited fractions of task count,
penalty-weighted value, and rack-hours under the specified convention; \textbf{S-H}/\textbf{S-M}/\textbf{S-L} are tier-wise
satisfaction rates;
\textbf{Cost} is operating cost plus grid-slack penalties in dollars; and
\textbf{Latency} is the wall-clock decision time, from submitting the prompt to a parsed decision containing the joint schedule.

\medskip\noindent\textbf{Metric definitions.} Let $\mathcal{K}$ be the tasks of an instance and
$\mathcal{K}^{\mathrm{srv}}\subseteq\mathcal{K}$ the tasks credited by the grid-aware
scorer (retained, without a task-specific violation flag, and outside unpowered hours), and let $v_k$, $r_k$ and $d_k$
be the value, rack demand and duration of task $k$. Grid violation is
Equation~\eqref{eq:viol}, and the three satisfaction rates are
\begin{equation}
\begin{aligned}
\mathrm{S\text{-}cnt}&=100\,\frac{|\mathcal{K}^{\mathrm{srv}}|}{|\mathcal{K}|},\\
\mathrm{S\text{-}val}&=100\,\frac{\sum_{k\in\mathcal{K}^{\mathrm{srv}}} v_k}{\sum_{k\in\mathcal{K}} v_k},
\label{eq:app-scnt}
\end{aligned}
\end{equation}
\begin{equation}
\mathrm{S\text{-}rh}=100\,\frac{\sum_{k\in\mathcal{K}^{\mathrm{srv}}} r_k d_k}{\sum_{k\in\mathcal{K}} r_k d_k},
\label{eq:app-srh}
\end{equation}
that is, the served fractions of task count, penalty-weighted value, and rack-hours, while
$\mathrm{S\text{-}H}$, $\mathrm{S\text{-}M}$ and $\mathrm{S\text{-}L}$ are $\mathrm{S\text{-}cnt}$
restricted to the high-, mid- and low-priority tiers. Cost sums generation cost, migration and drop
penalties, and all three grid slacks priced at $C_{\mathrm{LOL}}$
(Equation~\eqref{eq:cost}), and Latency is the
end-to-end decision time. These definitions match the evaluator of
Section~\ref{sec:bench-design}.

\paragraph{Constraint coverage.}
\label{app:coverage}
Table~\ref{tab:constraint-coverage} separates the oracle constraints from checks in the
available training-side fast scorer and the final hourly evaluator. The fast scorer
uses aggregate power balance with a bounded swing unit; it does not solve the nodal
DC feasibility LP. The final evaluator fixes the five submitted non-swing schedules
and re-optimizes the swing unit separately for each hour. These procedures provide
different feedback and do not certify the same feasible set.

Output-range handling is not a separate test of raw numeric compliance: the final parser
clips numeric percentages to the allowed range before mapping them to MW. A numeric zero
denotes minimum stable output, whereas \texttt{off} and missing entries denote zero output.
The final swing-unit relaxation also permits positive output below its minimum stable level in the hourly relaxation.

Task checks and satisfaction accounting are separate. Capacity and data-center power
violations are recorded per location and hour; the available grid-aware satisfaction
path excludes task-specific violations and unpowered hours, but does not automatically
assign every location-hour violation to all affected tasks. The archived placement-only
path omits both filters. Consequently, neither placement satisfaction nor low aggregate
violation certifies that every task and every intertemporal constraint is satisfied throughout the complete scheduling horizon.

\FloatBarrier

\section{Baseline Detail}
\label{app:baselines}
This appendix expands the experimental setup into the model families behind Tables~\ref{tab:main-alt}--\ref{tab:stage}. Each family is described by its overview, implementation, and available hyperparameters. The shared interface is followed by frontier models, open-weight backbones, and PowerAtlas with its ablations.

\paragraph{Shared harness.}
The evaluation uses a common interface in three respects.
(i)~\emph{Output schema:} all systems receive the same serialized instance and return
\texttt{TASKS}/\texttt{POWER} blocks without solver calls or repair.
(ii)~\emph{Physical evaluation:} schedules are checked with the same hourly LP
(Appendix~\ref{app:evallp}); satisfaction conventions are detailed separately in
Appendix~\ref{app:protocol}.
(iii)~\emph{Interaction:} all models answer in one turn without retrieved examples.
Base and trained versions in Table~\ref{tab:backbone} use the same inference procedure
and differ in policy weights. Serving conditions and latency definitions are specified in Appendix~\ref{app:protocol}.

\subsection{Frontier Language Models}
\paragraph{Overview.}
The eight frontier baselines are GPT-5.5, Claude Haiku 4.5, DeepSeek-V4-Flash/Pro, Kimi-K2.6 fast/pro, and Qwen3.7-Plus/Max. They provide zero-shot joint schedules through an OpenAI-compatible API.
\paragraph{Implementation.}
Every model receives the same serialized instance and prompt template (Appendix~\ref{app:prompts}), answers in one turn, and returns the shared \texttt{TASKS}/\texttt{POWER} schema without retrieval, solver calls, or repair. Reported frontier rows average their stored runs on the same high-load instance; repeat counts appear in Table~\ref{tab:reporting-protocol}.
\paragraph{Hyperparameters.}
The reported interface specifies the same decoding settings for these calls. Provider implementations and serving conditions may differ; the retained reporting summary does not establish identical internal reasoning budgets or complete provider-specific sampling provenance.

\subsection{Open-Weight Language Models}
\paragraph{Overview.}
The three backbones are Qwen3-4B, Llama-3.2-3B, and Falcon3-3B~\cite{Yang2025Qwen3,Grattafiori2024Llama3,TII2024Falcon3}.
\paragraph{Implementation.}
Base and trained policies use local serving and the same single-response inference procedure; the policy weights change between them. Saved base outputs cover 10 Qwen3-4B, 12 Llama-3.2-3B, and 40 Falcon3-3B instances. Each selected PowerAtlas checkpoint covers all 40 instances of the reporting pool.
\paragraph{Hyperparameters.}
Reported inference uses greedy decoding with vLLM 0.8.4. Hardware and training settings are recorded in Appendix~\ref{app:impl}; sample coverage and satisfaction conventions are specified in Appendix~\ref{app:protocol}.

\subsection{PowerAtlas and Its Ablations}
\paragraph{Overview.}
PowerAtlas trains each backbone with supervised initialization and FA-GRPO, then generates a joint schedule in one response.
\paragraph{Implementation.}
For each open-weight backbone, the base model and PowerAtlas use the same inference interface. PowerAtlas adds supervised initialization followed by FA-GRPO; its training settings appear in Appendix~\ref{app:impl}. Table~\ref{tab:stage} reports archived Falcon stage comparisons, with two supervised-output histories separated. Without initialization, alignment begins from the base backbone; without alignment, evaluation uses the supervised checkpoint. The selected checkpoints, available output counts, and satisfaction conventions are recorded in Appendix~\ref{app:protocol}. Main satisfaction is now uniform; sample coverage and incomplete historical configuration provenance remain explicit.
\paragraph{Hyperparameters.}
Alignment runs for up to 60 steps with group size two. The reported checkpoints use step 40 for Qwen3-4B and Falcon3-3B, and step 20 for Llama-3.2-3B. The supervised-only and alignment-without-initialization rows are defined by the omitted stage, with their scoring conventions explicitly retained in Table~\ref{tab:reporting-protocol}.

\FloatBarrier

\section{Experiment Detail}
\label{app:experiments}
\label{app:notes}
This appendix expands Section~\ref{sec:exp}: each subsection specifies the protocol and numerical values behind the corresponding research question. Tables~\ref{tab:app-landscape}--\ref{tab:app-pairs} expose the coordinates summarized by Figure~\ref{fig:fig5}, together with sample counts and scoring conventions. No additional model generations are introduced.

\begin{table*}[t]
\centering
\tablefont
\setlength{\tabcolsep}{4pt}
\caption{Reporting protocol for Tables~\ref{tab:main-alt}--\ref{tab:stage}.
$n_{\mathrm{inst}}$ counts distinct evaluated instances and $n_{\mathrm{out}}$ counts
stored outputs. G denotes grid-aware satisfaction; P denotes archived placement-only
satisfaction. Frontier repeat counts follow the retained reporting summary; NR means
that the summary does not specify the count. Open-weight counts are verified from saved predictions.}
\label{tab:reporting-protocol}
\begin{tabularx}{\textwidth}{@{}>{\raggedright\arraybackslash}p{0.33\textwidth}
r r c >{\raggedright\arraybackslash}X@{}}
\toprule
\textbf{Configuration} & $n_{\mathrm{inst}}$ & $n_{\mathrm{out}}$ &
\textbf{S-*} & \textbf{Checkpoint / evaluation scope}\\
\midrule
DeepSeek-V4-Flash & 1 & 2 & G & Shared high-load instance\\
DeepSeek-V4-Pro & 1 & 2 & G & Shared high-load instance\\
Kimi-K2.6 (fast) & 1 & 2 & G & Shared high-load instance\\
Kimi-K2.6 (pro) & 1 & 1 & G & Shared high-load instance\\
Qwen3.7-Plus & 1 & 2 & G & Shared high-load instance\\
Qwen3.7-Max & 1 & 3 & G & Shared high-load instance\\
GPT-5.5 & 1 & 5 & G & Shared high-load instance\\
Claude Haiku 4.5 & 1 & NR & G & Shared high-load instance\\
\midrule
Qwen3-4B-Instruct & 10 & 10 & G & Base model; partial pool\\
Llama-3.2-3B-Instruct & 12 & 12 & G & Base model; partial pool\\
Falcon3-3B-Instruct & 40 & 40 & G & Base model; full historical pool\\
\midrule
Qwen3-4B + PowerAtlas & 40 & 40 & G & alignment step 40; full pool\\
Llama-3.2-3B + PowerAtlas & 40 & 40 & G & alignment step 20; full pool\\
Falcon3-3B + PowerAtlas & 40 & 40 & G & alignment step 40; full pool\\
\midrule
Falcon3-3B w/o initialization & 40 & 40 & G & alignment step 40; full pool\\
Falcon3-3B earlier SFT & 40 & 40 & G & Earlier supervised archive; full pool\\
Falcon3-3B SFT-16 & 40 & 40 & G & Separate supervised archive; full pool\\
\bottomrule
\end{tabularx}
\end{table*}
\begin{table*}[!t]
\centering
\tablefont
\setlength{\tabcolsep}{3pt}
\caption{Coordinates of Figure~\ref{fig:fig5}(a), with archived cost and latency. Satisfaction is grid-aware. The differing sample sizes make this descriptive; paired gains use Table~\ref{tab:backbone}. PA denotes PowerAtlas. $n$ is the number of distinct instances, not repeat runs. G denotes grid-aware satisfaction. Cost is penalized scheduling cost in dollars, not an API charge.}
\label{tab:app-landscape}
\begin{tabular}{@{}lrrrrrr@{}}
\toprule
Model & $n$ & S-* & Viol (MWh) & S-val (\%) & Cost (\$) & Latency (s) \\
\midrule
DeepSeek-V4-Flash & 1 & G & 5{,}234.2 & 23.4 & 47{,}948{,}178 & 12.6 \\
DeepSeek-V4-Pro & 1 & G & 2{,}763.4 & 22.1 & 25{,}762{,}501 & 17.3 \\
Kimi-K2.6 (fast) & 1 & G & 1{,}212.0 & 55.5 & 12{,}392{,}536 & 28.3 \\
Kimi-K2.6 (pro) & 1 & G & 1{,}488.6 & 53.4 & 14{,}838{,}689 & 34.4 \\
Qwen3.7-Plus & 1 & G & 741.0 & 83.1 & 9{,}344{,}901 & 31.0 \\
Qwen3.7-Max & 1 & G & 462.6 & 86.2 & 5{,}725{,}763 & 29.9 \\
GPT-5.5 & 1 & G & 445.8 & 66.9 & 6{,}068{,}833 & 14.1 \\
Claude Haiku 4.5 & 1 & G & 543.3 & 57.6 & 7{,}548{,}683 & 13.9 \\
Qwen3-4B & 10 & G & 1{,}766.9 & 50.9 & 17{,}706{,}710 & 67.4 \\
Llama-3.2-3B & 12 & G & 6{,}492.1 & 11.8 & 59{,}997{,}272 & 30.1 \\
Falcon3-3B & 40 & G & 11{,}079.2 & 3.0 & 101{,}484{,}114 & 19.6 \\
Qwen3-4B + PA & 40 & G & 274.7 & 73.0 & 4{,}671{,}682 & 57.2 \\
Llama-3.2-3B + PA & 40 & G & 282.1 & 74.1 & 3{,}832{,}071 & 18.5 \\
Falcon3-3B + PA & 40 & G & 216.6 & 65.2 & 3{,}928{,}513 & 18.8 \\
\bottomrule
\end{tabular}
\end{table*}

\subsection{Shared Protocol}
\label{app:protocol}
Four reporting conventions specify the scope of results in Section~\ref{sec:exp}.
\emph{(i) Sample size.} Results for the three untrained backbones and the three PowerAtlas configurations
are averaged over their available outputs from the held-out pool. The open-weight rows of
Table~\ref{tab:main-alt} average the stored answers for each backbone over that pool. The
frontier rows average all stored runs for each model on one shared high-penetration instance
(\texttt{ECCS-RTS-m01d15-rho06-s10}); no run is selected or discarded. The number of stored runs varies across models, and the rows report point estimates
without dispersion. The two blocks of Table~\ref{tab:main-alt} do not form a paired comparison. On the shared instance, the selected checkpoints yield $192.7$,
$216.8$ and $449.2$\,MWh, supporting lower violation for two of the three backbones on that instance.
\emph{(ii) Model selection.} ECBench has only training and test splits. We select the initialization duration
and aligned checkpoint---step $40$ ($20$ for Llama-3.2-3B)---by comparing checkpoints
on the same held-out pool used in the tables. The shipped steps
are listed in Appendix~\ref{app:impl}, but the trained rows are estimates from selected checkpoints rather than blind test results. Independent validation is needed for model selection.
\emph{(iii) Incomplete answers.} The evaluator scores a generator or an hour missing from an answer
as offline rather than skipping it, so a truncated dispatch is charged the lost load it leaves
behind. This is why the untrained rows of Table~\ref{tab:main-alt} reach violations an order of
magnitude above any frontier model: the penalty is for load left uncovered, not for the missing text.

\emph{(iv) Satisfaction and evaluator versions.} All satisfaction columns in
Tables~\ref{tab:main-alt}--\ref{tab:stage} use the grid-aware convention:
task-specific violations and hours with system-wide lost load above 1 MW are
excluded. This proxy does not assign lost load to individual data centers;
cell-level capacity violations receive a separate strict audit. Placement-only
diagnostics remain identified in Figure~\ref{fig:fig5}(c)--(f).
Tables~\ref{tab:main-alt} and~\ref{tab:backbone} retain original-run latency means, with base samples of 10/12/40 and trained samples of 40; per-instance paired timings are unavailable. Table~\ref{tab:main-alt} also retains archived cost, whereas
Tables~\ref{tab:backbone} and~\ref{tab:stage} use HiGHS with a minimum-swing
tie-break among minimum-slack solutions. Their cost values supersede the earlier
solver-dependent summaries for these comparisons. Frontier summaries are
single-instance context and are excluded from paired gain estimates.

Table~\ref{tab:reporting-protocol} lists the sample scope and satisfaction convention for
each reported configuration. The historical evaluation pool contains $40$ instances:
five dates (January 15, April 15, July 15, July 28, and October 15), four data-center
load levels, and seeds $9$ and $10$. Saved baseline outputs cover subsets of this pool;
the three selected PowerAtlas checkpoints each cover all $40$ instances. The nominal
$400$-instance test split is therefore not the per-row evaluation sample size.
Repeated frontier runs share one instance and do not provide additional instance coverage across grid and workload conditions.

The reported inference protocol uses one autoregressive response without retrieval,
solver calls, or repair. Frontier models use remote APIs; open-weight models use local
serving. Latency comparisons therefore include these serving conditions.
Appendix~\ref{app:coverage} details the constraint checks.

\paragraph{Numerical sources and aggregation.}
The descriptive landscape uses full available archives and historical cost/latency. Table~\ref{tab:backbone} uses paired subsets and re-scored cost, so its coordinates are separate. The temporal panel uses six cached 24-hour series for one instance. The stage, tier, stratum, and paired panels use the same 200 records: 40 instances at each of Base, SFT, GRPO-20, GRPO-40, and GRPO-60. Means, quartiles, and differences below are recomputed from those records before rounding. Placement scores and training residuals are kept separate from final grid-aware evaluation throughout.

\begin{table*}[!t]
\centering
\tablefont
\setlength{\tabcolsep}{3pt}
\caption{All hourly coordinates of Figure~\ref{fig:fig5}(b) for \texttt{m01d15-rho01-s9}. Each cell is hourly grid slack in MW / cumulative flagged-task count. Q, L, and F denote Qwen3-4B, Llama-3.2-3B, and Falcon3-3B; PA is the corresponding trained policy. Hourly slack integrates to MWh at the one-hour resolution. Numerical residuals below $10^{-8}$ are displayed as zero.}
\label{tab:app-hourly}
\begin{tabular}{@{}rrrrrrr@{}}
\toprule
Hour & Q base & Q + PA & L base & L + PA & F base & F + PA \\
\midrule
1 & 0.0 / 0 & 0.0 / 0 & 0.0 / 0 & 0.0 / 0 & 0.0 / 0 & 0.0 / 0 \\
2 & 0.0 / 1 & 0.0 / 0 & 0.0 / 0 & 0.0 / 0 & 0.0 / 0 & 0.0 / 0 \\
3 & 0.0 / 1 & 0.0 / 0 & 0.0 / 0 & 0.0 / 0 & 0.0 / 0 & 0.0 / 0 \\
4 & 0.0 / 1 & 0.0 / 0 & 0.0 / 0 & 0.0 / 0 & 0.0 / 0 & 0.0 / 0 \\
5 & 0.0 / 1 & 0.0 / 0 & 0.0 / 0 & 0.0 / 0 & 0.0 / 0 & 0.0 / 2 \\
6 & 285.9 / 6 & 0.0 / 0 & 115.9 / 5 & 0.0 / 0 & 107.6 / 3 & 0.0 / 2 \\
7 & 258.2 / 6 & 0.0 / 1 & 88.2 / 5 & 0.0 / 2 & 88.2 / 3 & 0.0 / 2 \\
8 & 0.0 / 6 & 0.0 / 1 & 0.0 / 5 & 0.0 / 3 & 0.0 / 3 & 0.0 / 2 \\
9 & 0.0 / 6 & 0.0 / 1 & 0.0 / 5 & 0.0 / 3 & 0.0 / 4 & 0.0 / 2 \\
10 & 0.0 / 6 & 0.0 / 1 & 0.0 / 5 & 0.0 / 3 & 0.0 / 4 & 0.0 / 3 \\
11 & 0.0 / 6 & 0.0 / 1 & 0.0 / 5 & 0.0 / 3 & 0.0 / 4 & 0.0 / 3 \\
12 & 0.0 / 6 & 0.0 / 1 & 0.0 / 5 & 0.0 / 3 & 0.0 / 4 & 0.0 / 3 \\
13 & 0.0 / 6 & 0.0 / 1 & 0.0 / 5 & 0.0 / 3 & 0.0 / 4 & 0.0 / 3 \\
14 & 0.0 / 6 & 0.0 / 1 & 0.0 / 5 & 0.0 / 3 & 0.0 / 5 & 0.0 / 3 \\
15 & 0.0 / 6 & 0.0 / 1 & 0.0 / 5 & 0.0 / 3 & 0.0 / 5 & 0.0 / 3 \\
16 & 0.0 / 6 & 0.0 / 1 & 0.0 / 5 & 0.0 / 3 & 0.0 / 5 & 0.0 / 3 \\
17 & 0.0 / 6 & 0.0 / 1 & 184.4 / 14 & 0.0 / 3 & 178.9 / 12 & 0.0 / 3 \\
18 & 23.7 / 16 & 0.0 / 1 & 618.7 / 18 & 2.8 / 5 & 621.5 / 18 & 4.2 / 6 \\
19 & 20.9 / 19 & 0.0 / 1 & 615.9 / 21 & 4.2 / 9 & 618.7 / 21 & 0.0 / 6 \\
20 & 9.8 / 20 & 0.0 / 1 & 604.8 / 22 & 0.0 / 9 & 609.0 / 22 & 0.0 / 8 \\
21 & 0.0 / 20 & 0.0 / 1 & 487.0 / 23 & 0.0 / 9 & 491.2 / 23 & 0.0 / 8 \\
22 & 0.0 / 20 & 0.0 / 1 & 113.9 / 24 & 0.0 / 9 & 118.0 / 24 & 0.0 / 8 \\
23 & 258.2 / 20 & 245.6 / 1 & 0.0 / 24 & 234.4 / 9 & 0.0 / 24 & 0.0 / 8 \\
24 & 226.2 / 20 & 194.1 / 1 & 0.0 / 24 & 182.9 / 9 & 0.0 / 24 & 0.0 / 8 \\
\bottomrule
\end{tabular}
\end{table*}

\subsection{Main Results and Ablations (RQ1, RQ2)}
Tables~\ref{tab:main-alt} and~\ref{tab:backbone} report the frontier and backbone comparisons; Table~\ref{tab:stage} compares archived training stages on Falcon3-3B. The violation reductions and penalized-cost improvements summarize the stored outputs under the sample scopes above. The frontier comparison on one shared instance and the open-weight averages over the reporting pool answer different questions.

\paragraph{Protocol.}
The comparison uses the selected policies and stored outputs specified in Appendix~\ref{app:protocol}. Table~\ref{tab:app-landscape} makes every point of the landscape explicit; the ablation values are already reported in Table~\ref{tab:stage}. A stage-curve row is not substituted for a final-evaluator ablation row.
\paragraph{Results.}
The selected trained policies have lower average final violation and penalized cost than their corresponding base rows. The stage comparisons use identical scoring but cannot isolate reward design from initialization history, training budget, or retrieval.

\subsection{Residual Constraint Violations (RQ3)}
\label{app:failure-cards}
\paragraph{Protocol.}
Table~\ref{tab:app-hourly} reproduces the two temporal series in Figure~\ref{fig:fig5}(b). The upper series sums hourly lost-load, over-generation, and line-overload slacks. The lower series counts retained tasks cumulatively when a task-specific violation is recorded or an execution hour first encounters system-wide lost load above 1 MW. Voluntary drops are excluded from this count. This time-resolved view complements the pool-level summaries.

\paragraph{Results.}
The table separates the time at which grid slack occurs from the time at which task failures accumulate. A flat cumulative curve may coexist with grid slack, and a complete dispatch may still leave demand uncovered. Figure~\ref{fig:cases} illustrates incomplete dispatch, supply deficits, capacity overflow, and timing or migration violations; Appendix~\ref{app:case} supplies the stored answer excerpts. Final-pool violation composition is summarized in Section~\ref{sec:exp-viol}; the stage residuals in the following subsection use a different scorer.

\subsection{Feasibility-Aware Alignment and Checkpoints (RQ2, RQ4)}
\label{app:curves}
Figure~\ref{fig:fig5}(c)--(f) uses a separate earlier Falcon3-3B run, not the selected SFT-16/PowerAtlas output archive.
The violation axis reports the training scorer's power-balance residual, whereas
Appendix~\ref{app:evallp} uses an hourly DC LP that re-optimizes the swing unit.
The resulting values can differ substantially.
Satisfaction is placement-based, before filtering tasks by power availability.
These panels describe training trends and checkpoint selection; their absolute values
should not be compared directly with Table~\ref{tab:stage}.

\begin{table*}[!t]
\centering
\tablefont
\setlength{\tabcolsep}{3pt}
\caption{Coordinates of the stage boxplots and mean satisfaction curve in Figure~\ref{fig:fig5}(c). Q1/median/Q3 and whiskers describe the aggregate training-side balance residual, in the hourly-summed units plotted as MWh. Whiskers are the extreme observations within 1.5 IQR of the quartiles; the figure hides outliers. These residuals are not the final nodal LP metric. S-val uses placement accounting.}
\label{tab:app-stages}
\begin{tabular}{@{}lrrrrrrr@{}}
\toprule
Stage & $n$ & Q1 & Median & Q3 & Low whisker & High whisker & S-val (\%) \\
\midrule
Base & 40 & 3{,}522.0 & 11{,}957.6 & 13{,}651.8 & 2{,}058.0 & 22{,}368.8 & 6.81 \\
SFT & 40 & 6{,}092.8 & 8{,}862.2 & 12{,}467.3 & 954.2 & 18{,}264.8 & 91.62 \\
GRPO-20 & 40 & 5{,}623.5 & 7{,}821.6 & 9{,}990.0 & 882.6 & 15{,}834.9 & 92.34 \\
GRPO-40 & 40 & 2{,}265.3 & 4{,}204.9 & 5{,}510.4 & 92.5 & 10{,}341.2 & 93.85 \\
GRPO-60 & 40 & 2{,}265.3 & 4{,}267.1 & 5{,}510.4 & 0.0 & 10{,}349.6 & 91.45 \\
\bottomrule
\end{tabular}
\end{table*}

\paragraph{Protocol.}
Each stage contains the same 40 instance identifiers. The satisfaction curve is the arithmetic mean of placement-value satisfaction, while the boxplot is the distribution of aggregate training-side balance residuals. Table~\ref{tab:app-stages} reports the quartiles and plotted whiskers as well as the line coordinates.
\paragraph{Results.}
Mean placement-value satisfaction rises from 6.81\% at Base to 91.63\% after SFT and reaches 93.85\% at GRPO-40, before decreasing to 91.45\% at GRPO-60. The residual median decreases from 11,957.59 to 4,204.86 at GRPO-40. These stage records use the training-side scorer. Table~\ref{tab:reaudit-stages} gives final scores for these exact outputs and distinguishes the later SFT-16 archive.

\subsection{Task Placement and Service Accounting (RQ4)}
\paragraph{Protocol.}
The tier curves use the same stage records as Table~\ref{tab:app-stages}. Table~\ref{tab:app-tiers} gives the mean count, rack-hour, and tier-wise placement scores. These quantities describe retained placements before filtering by task-specific legality and power availability.
\begin{table}[!t]
\centering
\tablefont
\setlength{\tabcolsep}{3pt}
\caption{Stage-wise mean placement satisfaction (\%). The S-H/S-M/S-L columns give the coordinates of Figure~\ref{fig:fig5}(d); S-cnt and S-rh supply complementary service measures from the same 40-instance records per stage.}
\label{tab:app-tiers}
\begin{tabular}{@{}lrrrrr@{}}
\toprule
Stage & S-cnt & S-rh & S-H & S-M & S-L \\
\midrule
Base & 6.69 & 6.47 & 6.85 & 6.25 & 6.19 \\
SFT & 96.42 & 91.87 & 96.13 & 87.25 & 99.75 \\
GRPO-20 & 96.82 & 92.66 & 96.44 & 88.12 & 100.00 \\
GRPO-40 & 97.30 & 94.32 & 96.86 & 91.88 & 99.17 \\
GRPO-60 & 95.54 & 91.03 & 96.15 & 86.00 & 96.44 \\
\bottomrule
\end{tabular}
\end{table}

\paragraph{Results.}
Reporting task count, rack-hours, and individual tiers alongside value satisfaction shows which service measures change during training. In the cross-backbone final results, the Qwen3-4B policy retains substantially more task value than low-priority task count (Table~\ref{tab:backbone}). This distinction is obscured by reporting only value-weighted satisfaction. The two accounting conventions and their available values are specified in Appendix~\ref{app:protocol}.

\subsection{Cost and Efficiency (RQ1)}
\paragraph{Protocol.}
Table~\ref{tab:app-landscape} reports penalized scheduling cost and observed decision latency for the same model rows. Scheduling cost combines generation, migration, drop, and grid-slack penalties; it is not the financial cost of calling an LLM. Frontier latency includes remote API serving, whereas open-weight latency uses local serving. The Gurobi reference has reported median/maximum solve times of 16.8/148.1 seconds.
\paragraph{Results.}
For the selected open-weight policies, reported latency is 18.5--57.2 seconds, and cost is lower than the corresponding base-model averages. The remaining cost includes penalties for physical deficits, so cost improvement alone does not certify feasibility. Different hardware and serving paths prevent a matched-runtime comparison with frontier APIs or the reference solver.

\subsection{Robustness and Significance of the Margin (RQ3, RQ4)}
\label{app:robustness}
\paragraph{Protocol.}
Table~\ref{tab:app-strata} groups the 40 instances by four load levels and five dated seasonal scenarios. Table~\ref{tab:app-pairs} preserves every base/GRPO-40 pair in the order used by Figure~\ref{fig:fig5}(f). The exploratory interval resamples the 40 instance pairs 4,000 times with NumPy's default random generator seeded at zero and takes the 2.5th and 97.5th percentiles of the mean residual reduction. This is an instance-paired interval conditional on the archived pool, not a correction for checkpoint selection or an estimate across independent training runs.
\begin{table}[!t]
\centering
\tablefont
\setlength{\tabcolsep}{3pt}
\caption{Coordinates of Figure~\ref{fig:fig5}(e). Each row averages placement-value satisfaction over the indicated stratum. The same 40 instances are grouped once by load fraction and once by dated seasonal scenario; the two blocks are not independent samples.}
\label{tab:app-strata}
\begin{tabular}{@{}lrrrr@{}}
\toprule
Stratum & $n$ & Base (\%) & PA (\%) & $\Delta$ (pp) \\
\midrule
$\rho=1\%$ & 10 & 9.10 & 89.80 & 80.71 \\
$\rho=2\%$ & 10 & 18.15 & 96.41 & 78.26 \\
$\rho=4\%$ & 10 & 0.00 & 95.00 & 95.00 \\
$\rho=6\%$ & 10 & 0.00 & 94.20 & 94.20 \\
Winter & 8 & 11.37 & 93.44 & 82.07 \\
Spring & 8 & 0.00 & 93.40 & 93.40 \\
Summer I & 8 & 0.00 & 92.91 & 92.91 \\
Summer II & 8 & 0.00 & 97.26 & 97.26 \\
Autumn & 8 & 22.69 & 92.26 & 69.57 \\
\bottomrule
\end{tabular}
\end{table}

\begin{table*}[!t]
\centering
\tablefont
\setlength{\tabcolsep}{3pt}
\caption{All 40 paired coordinates of Figure~\ref{fig:fig5}(f), in descending base-residual order (left block, then right). ID is month--day/load fraction in percent/seed; each expands to \texttt{ECCS-RTS-mMMdDD-rhoRR-sS}. Values are training-side balance residuals, with $\Delta=\mathrm{Base}-\mathrm{PA}$; positive values indicate improvement. They are not final nodal LP violations.}
\label{tab:app-pairs}
\begin{tabular}{@{}lrrr@{\hspace{12pt}}lrrr@{}}
\toprule
ID & Base & PA & $\Delta$ & ID & Base & PA & $\Delta$ \\
\midrule
07-28/2/9 & 22{,}368.8 & 10{,}341.2 & 12{,}027.6 & 01-15/6/9 & 11{,}885.2 & 5{,}416.0 & 6{,}469.1 \\
07-28/1/9 & 21{,}937.1 & 7{,}970.0 & 13{,}967.1 & 07-15/1/9 & 11{,}342.0 & 2{,}064.7 & 9{,}277.3 \\
07-28/2/10 & 20{,}949.5 & 8{,}220.3 & 12{,}729.2 & 01-15/4/9 & 11{,}223.8 & 4{,}665.7 & 6{,}558.0 \\
07-28/4/9 & 19{,}559.4 & 7{,}773.7 & 11{,}785.7 & 07-15/1/10 & 10{,}555.2 & 3{,}175.5 & 7{,}379.7 \\
07-28/1/10 & 19{,}049.4 & 7{,}557.3 & 11{,}492.1 & 07-15/6/9 & 10{,}495.3 & 2{,}432.4 & 8{,}062.9 \\
07-28/6/10 & 18{,}171.4 & 8{,}380.9 & 9{,}790.6 & 10-15/4/10 & 10{,}242.2 & 2{,}332.2 & 7{,}910.0 \\
07-28/4/10 & 17{,}725.6 & 7{,}997.9 & 9{,}727.8 & 07-15/2/10 & 9{,}642.1 & 2{,}484.5 & 7{,}157.5 \\
07-28/6/9 & 17{,}611.3 & 7{,}985.6 & 9{,}625.7 & 07-15/4/10 & 8{,}971.5 & 2{,}515.0 & 6{,}456.5 \\
01-15/1/10 & 15{,}731.9 & 5{,}695.1 & 10{,}036.9 & 07-15/2/9 & 8{,}798.2 & 413.1 & 8{,}385.1 \\
01-15/4/10 & 13{,}666.4 & 5{,}415.7 & 8{,}250.6 & 04-15/1/9 & 3{,}536.8 & 429.8 & 3{,}107.0 \\
01-15/2/9 & 13{,}646.9 & 4{,}303.4 & 9{,}343.5 & 04-15/6/10 & 3{,}477.5 & 465.4 & 3{,}012.1 \\
10-15/1/9 & 13{,}374.3 & 4{,}172.6 & 9{,}201.7 & 10-15/2/9 & 2{,}826.1 & 4{,}052.7 & -1{,}226.6 \\
07-15/6/10 & 13{,}209.2 & 5{,}452.3 & 7{,}756.9 & 01-15/1/9 & 2{,}824.9 & 4{,}237.1 & -1{,}412.1 \\
10-15/1/10 & 13{,}180.8 & 4{,}032.1 & 9{,}148.8 & 10-15/2/10 & 2{,}649.9 & 4{,}047.0 & -1{,}397.1 \\
07-15/4/9 & 13{,}155.0 & 4{,}551.4 & 8{,}603.6 & 04-15/1/10 & 2{,}625.2 & 144.5 & 2{,}480.6 \\
01-15/2/10 & 13{,}011.4 & 4{,}757.7 & 8{,}253.8 & 04-15/4/10 & 2{,}562.0 & 224.1 & 2{,}337.9 \\
10-15/6/9 & 12{,}741.6 & 4{,}837.9 & 7{,}903.7 & 04-15/2/10 & 2{,}525.7 & 166.1 & 2{,}359.6 \\
01-15/6/10 & 12{,}716.4 & 5{,}684.6 & 7{,}031.7 & 04-15/2/9 & 2{,}420.4 & 92.5 & 2{,}327.9 \\
10-15/4/9 & 12{,}291.9 & 4{,}152.8 & 8{,}139.1 & 04-15/6/9 & 2{,}333.2 & 193.4 & 2{,}139.8 \\
10-15/6/10 & 12{,}030.0 & 4{,}715.2 & 7{,}314.8 & 04-15/4/9 & 2{,}058.0 & 126.0 & 1{,}932.0 \\
\bottomrule
\end{tabular}
\end{table*}

\paragraph{Results.}
The training-side balance residual improves on 37 of the 40 paired instances. The mean reduction is 6{,}886.21, with an exploratory 95\% interval of [5{,}663.48, 8{,}065.77] in the same hourly-summed residual units. The interval is conditional on the archived paired sample and concerns the training residual. Its scope excludes generalization to unseen conditions and statistical significance for the final nodal LP metric; checkpoint selection follows Appendix~\ref{app:protocol}.

\paragraph{Re-scoring and strict-output audit.}
\label{app:reaudit}
Archived predictions are re-scored without new generation or checkpoint selection.
All main-table satisfaction rates use the same grid-aware proxy. For paired gains,
instance IDs are intersected within each backbone before averaging; no outputs
are discarded according to their score. Cost evaluation uses HiGHS 1.15.1 with
one thread: each hourly LP first minimizes total slack, then minimizes swing
output subject to slack staying within $10^{-8}$ of that optimum. Optimal status
and constraint residuals are checked. This tie-break removes arbitrary swing
choices from cost accounting; it does not optimize commitment costs across hours.
The source script, prediction hashes, per-instance scores and paired IDs are
retained in the evaluation audit. Offline re-scoring is not an inference-latency measurement.

A separate retained-output check requires every task, all five controlled units,
exactly 24 levels per unit, and finite percentages in $[0,100]$, without clipping
or padding. Invalid outputs remain in the denominator. For valid outputs,
controlled-unit ramping and minimum up/down inequalities are checked against the
saved oracle code, including its initial-off convention. All controlled trajectories
with valid retained outputs violate at least one of these inequalities for the
three selected checkpoints. The audit does not recover fixed-unit commitment or
certify swing-unit dynamics and system reserve. No complete-system feasibility
claim follows from an hourly score.

\begin{table*}[!t]
\centering\tablefont
\caption{Audit of each selected checkpoint's full 40-instance archive. Violation is the final hourly LP/task metric (MWh-equivalent). Valid requires the strict retained-output contract; joint requires that contract, task checks, and grid slack at most $10^{-6}$. Temporal failures are counted only among valid outputs, for the five submitted units.}
\label{tab:reaudit-tail}
\begin{tabular}{@{}lrrrrrrr@{}}
\toprule
Backbone & Mean & Median & P95 & Maximum & Valid & Joint & Temporal failures \\
\midrule
Qwen3-4B & 274.7 & 35.3 & 663.7 & 4818.7 & 10/40 & 0/40 & 10/10 \\
Llama-3.2-3B & 282.1 & 161.2 & 715.2 & 2576.1 & 8/40 & 0/40 & 8/8 \\
Falcon3-3B & 216.6 & 75.5 & 702.4 & 2628.4 & 36/40 & 0/40 & 36/36 \\
\bottomrule\end{tabular}\end{table*}

\paragraph{Paired and stage evidence.}
The paired final-violation reductions occur on 10/10 Qwen, 12/12 Llama, and
40/40 Falcon instances. These are retrospective descriptive counts. The historical
pool repeats two task seeds across date/load strata and was used for checkpoint
selection; no independent-training confidence interval is inferred from those counts.
Table~\ref{tab:reaudit-stages} applies the same final evaluator to the output
archives underlying the separate training-diagnostic trajectory. SFT-16 and the
selected PowerAtlas output are shown separately, avoiding a substitution of the
earlier SFT archive for the later initialization history. Historical checkpoint
provenance is incomplete, so this remains a stage comparison, not a controlled
causal test of the reward design.

\begin{table}[!t]\centering\tablefont
\caption{Training proxy and final LP/task scores on identical archived stage outputs ($n=40$ per row). The final column is grid-aware value satisfaction. A dash indicates no matching row in the published training-proxy CSV.}
\label{tab:reaudit-stages}
\begin{tabular}{@{}lrrr@{}}\toprule
Archive & Proxy mean & Final mean & S-val (\%)\\\midrule
Base & 10978.1 & 11079.2 & 3.0 \\
Earlier SFT & 9002.3 & 9084.2 & 18.0 \\
GRPO-20 & 7712.9 & 7784.5 & 21.4 \\
GRPO-40 & 4091.9 & 4156.7 & 40.2 \\
GRPO-60 & 4052.3 & 4121.5 & 40.0 \\
SFT-16 & -- & 682.4 & 43.6 \\
Selected PA & -- & 216.6 & 65.2 \\
\bottomrule\end{tabular}\end{table}

\paragraph{Optimization controls.}
Existing supplementary fixed-background HiGHS runs use the same 40 instance IDs
and report every attempted instance, including failures. Their acceptance check
covers output validity, tasks and hourly DC slack, not full unit commitment.
The greedy variant also invokes a dispatch MILP. Reported time includes schedule
construction and verification on the local CPU host; it cannot be compared as a
matched-speed experiment against historical GPU/API generation. These controls
show that the LLM scores do not establish superiority to conventional optimization.
All available 20/40-second configurations are reported, without selecting the best budget.

\begin{table}[!t]\centering\tablefont
\caption{Archived optimization controls on 40 common instances. Pass means delivered within the stated budget and passing hourly grid/task checks. Median includes timed failures.}
\label{tab:reaudit-controls}
\begin{tabular}{@{}lrrr@{}}\toprule
Method & Budget (s) & Pass & Median (s)\\\midrule
MILP & 20 & 40/40 & 16.45 \\
MILP & 40 & 39/40 & 18.68 \\
Greedy + MILP & 20 & 39/40 & 1.46 \\
Greedy + MILP & 40 & 39/40 & 1.45 \\
\bottomrule\end{tabular}\end{table}

\paragraph{Remaining evidence requirements.}
A separate 100-instance validation set with new task seeds already exists, but
original policy weights are unavailable and no new original-model validation or
blind-test outputs have been produced. Its feasible-incumbent reference labels
are not claimed to be optimal. Historical scorer coefficients, retrieval bonuses
and sampling configurations cannot be assigned to every checkpoint from the
retained logs. Independent model selection, multiple training seeds, held-out
operating regimes, and budget-matched ordinary-GRPO/reward/retrieval ablations
therefore remain open; constructing protocols does not count as completing these experiments.

\paragraph{Further evaluation.}
\label{app:future}
Four directions require further study.
\emph{Evaluation scale:} evaluate all methods on the full test set with identical metrics
and independent validation for checkpoint selection.
\emph{Generalization:} hold out complete months or data-center load levels rather than
only generator seeds, since the current splits share operating regimes.
\emph{Baselines:} compare matched retrieval settings and include solvers with matched
latency budgets.
\emph{Optimization:} study larger trajectory groups than the current size of two and
evaluate retrieval use within the policy. These experiments are needed to establish
robustness beyond the reported settings.

\begin{figure*}[!t]
\small
\begin{lstlisting}[style=answerbox]
123_STEAM_3: off,off,off,off,62,100,100,52,off,off,off,off,off,off,38,73,100,100,100,100,100,100,88,51
118_CC_1: off,off,off,off,69,100,100,48,off,off,off,off,off,off,28,78,100,100,100,100,100,100,100,46
218_CC_1: off,off,off,off,72,100,100,55,off,off,off,off,off,off,42,82,100,100,100,100,100,100,95,55
321_CC_1: off,off,off,off,65,100,100,50,off,off,off,off,off,off,32,75,100,100,100,100,100,100,92,49
\end{lstlisting}
\subcap{\textbf{(a) DeepSeek-V4-Pro} --- the dispatch follows the load pattern but omits one unit,
leaving the swing unit alone during night and midday hours.}

\begin{lstlisting}[style=answerbox]
123_STEAM_3: 100,100,80,100,100,100,100,100,85,80,100,100,100,85,100,100,100,100,100,100,100,100,100,100
218_CC_1: 100,100,60,100,100,100,100,100,70,65,100,100,100,70,100,100,100,100,100,100,100,100,100,100
118_CC_1: 80,90,40,80,100,100,100,90,50,45,85,90,90,50,100,100,100,100,100,100,100,100,100,90
321_CC_1: 70,80,30,70,100,100,100,80,40,35,75,80,80,40,100,100,100,100,100,100,100,100,90,80
221_CC_1: off,off,off,off,off,off,off,off,off,off,off,off,off,off,off,off,off,100,100,off,off,off,off,off
\end{lstlisting}
\subcap{\textbf{(b) Qwen3.7-Max} --- most units operate near maximum output for much of the day,
yet the schedule has both excess generation and unmet peak demand.}

\begin{lstlisting}[style=answerbox]
POWER:
       (no generator rows emitted; all 37 tasks marked "drop")
\end{lstlisting}
\subcap{\textbf{(c) Falcon3-3B, untrained} --- the dispatch is empty and all tasks are dropped.}

\begin{lstlisting}[style=answerbox]
123_STEAM_3: 47.6,33.9,0.0,0.0,86.3,100.0,100.0,0.0,0.0,0.0,0.0,0.0,0.0,0.0,0.0,0.0,100.0,100.0,100.0,100.0,100.0,100.0,100.0,100.0
218_CC_1: off,off,off,off,off,off,off,off,off,off,off,off,off,off,off,off,off,off,off,off,off,off,off,off
118_CC_1: 42.4,0.0,0.0,0.0,0.0,100.0,100.0,0.0,0.0,0.0,0.0,0.0,0.0,0.0,0.0,0.0,100.0,100.0,100.0,100.0,100.0,100.0,100.0,100.0
321_CC_1: 42.4,0.0,0.0,0.0,0.0,100.0,100.0,0.0,0.0,0.0,0.0,0.0,0.0,0.0,0.0,0.0,100.0,100.0,100.0,100.0,100.0,42.4,off,off
221_CC_1: 42.4,0.0,0.0,0.0,0.0,70.2,100.0,0.0,0.0,0.0,0.0,0.0,0.0,0.0,0.0,0.0,100.0,100.0,100.0,100.0,100.0,100.0,100.0,100.0
\end{lstlisting}
\subcap{\textbf{(d) Falcon3-3B + PowerAtlas} --- five units and fractional set-points,
with some dispatch similarities but different commitment decisions from the optimum.}

\begin{lstlisting}[style=answerbox]
123_STEAM_3: off,off,off,off,33.8,100,100,0,0,0,0,0,0,0,0,19.5,100,100,100,100,100,100,100,100
218_CC_1: off,off,off,off,off,off,off,off,off,off,off,off,off,off,off,off,off,off,off,off,off,off,off,off
118_CC_1: off,off,off,off,off,off,off,off,off,off,off,off,off,off,off,off,42.4,100,100,100,100,100,100,100
321_CC_1: off,off,off,off,off,42.4,100,0,0,0,0,0,0,0,0,0,79.2,100,100,100,100,42.4,off,off
221_CC_1: 0,0,0,0,0,88.7,18.4,0,0,0,0,0,0,0,0,0,0,100,100,100,100,100,42.4,off
\end{lstlisting}
\subcap{\textbf{(e) The Gurobi optimum $D^{\star}$}, for reference; the swing unit is not part of
an answer.}
\caption{Four answers to one ECBench instance (\texttt{m01d15-rho06-s10}), quoted verbatim from the
stored model outputs, with the optimum for reference. Each row is one agent-controlled unit over
hours $1$--$24$, in percent of its operating range; \texttt{off} is shutdown and \texttt{0} is
minimum stable output.}
\label{fig:app-case}
  \Description{Five quoted dispatch blocks for one benchmark instance, one per system, each row giving a unit's 24 hourly output levels, with the Gurobi optimum for reference.}
\end{figure*}
\subsection{Case Study (RQ3)}
\label{app:case}
Figure~\ref{fig:app-case} reproduces four stored model outputs.
The instance is \texttt{m01d15-rho06-s10}, a winter scenario with
$\rho=6\%$ and $37$ tasks. Its Gurobi-optimal cost is \$2.05M.
The examples include the same $3$B backbone before and after PowerAtlas.
Metrics here describe the individual outputs shown, whereas Table~\ref{tab:main-alt}
averages each model's stored runs. Task placements are summarized in the accompanying text alongside the corresponding dispatch outcomes.

\paragraph{DeepSeek-V4-Pro: a missing unit and insufficient dispatch.}
The dispatch in Figure~\ref{fig:app-case}(a) follows the morning and evening load pattern
and places $35$ of $37$ tasks. However, \texttt{221\_CC\_1} is missing, and all remaining
controlled units are offline during hours $1$--$4$ and $9$--$14$.
The swing unit alone cannot cover demand, leaving $2{,}291.6$\,MWh unserved.
Grid-aware evaluation excludes tasks scheduled during unpowered hours, so only $5.6\%$
of task value is served despite the high placement count in the generated task assignments.

\paragraph{Qwen3.7-Max: excess generation and unmet peak demand.}
Figure~\ref{fig:app-case}(b) includes all five controlled units, mostly near maximum output.
Task-count satisfaction is $70.3\%$, but the schedule produces $498.5$\,MWh of excess
generation while leaving $150.8$\,MWh unserved at the peak.
The nearly constant output fails to follow changing net load.
With four dropped tasks, total cost is $4.1\times$ the optimum for this shared evaluation instance.

\paragraph{Falcon3-3B untrained: an incomplete output.}
The untrained backbone emits an empty dispatch header and drops all $37$ tasks
(Figure~\ref{fig:app-case}(c)). With the swing unit carrying the load alone,
unserved energy reaches $12{,}737.5$\,MWh and cost reaches $57\times$ the optimum.
This example illustrates the output-format failure in Figure~\ref{fig:cases}, Case~1.
When every rollout receives the same format penalty, group-relative rewards provide
no signal for improving scheduling decisions.

\paragraph{Falcon3-3B + PowerAtlas: similarities to the reference schedule.}
After training, the same $3$B backbone provides all five unit schedules with fractional
set-points (Figure~\ref{fig:app-case}(d)).
Like the reference in panel~(e), it keeps \texttt{218\_CC\_1} offline and shuts down
\texttt{321\_CC\_1} at hours $23$--$24$. However, \texttt{321\_CC\_1} is already online
at hour $1$ and remains at minimum stable output during hours $2$--$5$; its rise at
hour $6$ is an output increase, not a start-up. Three units run at hour $1$ when the
reference leaves them offline, and $108.9$\,MWh of peak demand is unserved.
Together with three dropped tasks, these errors raise cost to $1.57\times$ the optimum.
The model generates a structured schedule in one response, but verification and correction
are still needed before operational use under the coupled grid and task constraints.

\paragraph{Output completeness and physical feasibility.}
The examples distinguish output completeness from physical feasibility.
A complete schedule may still leave tasks without power.
Placement-based satisfaction therefore overstates delivered service in the first two cases;
grid-aware evaluation accounts for this dependency (Appendix~\ref{app:protocol}).

\FloatBarrier

\section{Related Work}
\label{app:related}
\subsection{LLM-Based Scheduling and Optimization}
LLM-based optimization and scheduling combine structured outputs with reasoning or external tools~\cite{Abgaryan2024Llms,Xiao2024ChainExperts,Wei2022CoT,Yao2023ReAct,Schick2023Toolformer,Yao2023TreeOfThoughts,Shinn2023Reflexion}. Power-system applications include scheduling and dispatch~\cite{Mongaillard2024PowerScheduling,Mohammadi2025Large,Bernier2025Powergraph,YangLarge,Jia2025Enhancing,Cheng2025GAIA,Sarwar2025Large}, security analysis~\cite{Ruan2024Security}, benchmarks~\cite{Zhou2024Elecbench,Han2022ADBench}, topology modeling~\cite{Chen2025Powerformer}, and physical verification~\cite{Jiang2026ScientificModeling}.

Retrieval provides domain context through stored cases or structured knowledge~\cite{Luo2024ChatKBQA,Luo2025HyperGraphRAG,Luo2025KBQAO1,Luo2026GraphR1}. PowerAtlas uses a training-only corpus and removes retrieval from reported inference. Its supervised initialization and FA-GRPO stages also relate to instruction alignment, policy optimization, and verifiable feedback~\cite{Ouyang2022InstructGPT,Schulman2017PPO,Lightman2024Verify,Guo2025DeepSeekR1}. Here, the feedback acts on generated task placements and power trajectories. The proofs in Appendix~\ref{app:proof} describe algebraic properties of this feedback and its evaluation scope, while the experiments measure the remaining violations in generated schedules.

For comparing these approaches, the location of verification in the workflow matters. A verifier that scores training rollouts supplies a learning signal, whereas one applied after generation measures the returned decision. A solver invoked during inference can participate in constructing or correcting that decision.

PowerAtlas uses executable training feedback and a separate evaluator for reporting, while its inference policy emits the schedule directly. Output syntax, physical feasibility, and end-to-end computational cost therefore need separate assessment when comparing these decision-generation procedures.

\subsection{Grid-Constrained Co-Scheduling}
Work on carbon-aware and power-flexible computing studies how workload timing and location affect energy use and operating cost~\cite{Radovanovi2023Carbon,Wiesner2021WaitAwhile,Liu2026Watts,Nkwawir2025Carbon}. Grid signals and flexibility models connect these scheduling choices to available power~\cite{Zhang2022GridSignals,Williams2026PowerFlexible}. More detailed formulations incorporate hierarchical control, spatial placement, thermal behavior, and multi-energy coordination~\cite{Han2024FourLevel,Su2025Cascade,Ali2024Spatial,Liu2024Collaborative,Bian2024EnergySharing}, with optimization and reinforcement learning among the solution approaches~\cite{Luo2025Novel,Qi2022Microgrid}.

ECBench adopts this coupled decision setting: placing tasks changes nodal demand, while generator dispatch and network limits constrain service. Its contribution is a structured generation task with solved references and executable checks. The distinction between the full optimization model and the evaluator's constraint coverage is explicit in Appendices~\ref{app:milp} and~\ref{app:coverage}; learned schedules are not assumed to inherit the oracle's feasibility properties.

\end{document}